\documentclass[sigconf]{acmart}
\AtBeginDocument{%
  \providecommand\BibTeX{{%
    \normalfont B\kern-0.5em{\scshape i\kern-0.25em b}\kern-0.8em\TeX}}}

\copyrightyear{2021}
\acmYear{2021} 
\setcopyright{iw3c2w3}
\acmConference[WWW '21]{Proceedings of the Web Conference 2021}{April 19--23, 2021}{Ljubljana, Slovenia} 
\acmBooktitle{Proceedings of the Web Conference 2021 (WWW '21), April 19--23, 2021, Ljubljana, Slovenia}
\acmPrice{}
\acmDOI{10.1145/3442381.3450058}
\acmISBN{978-1-4503-8312-7/21/04}
\usepackage[utf8]{inputenc} % allow utf-8 input
\usepackage[T1]{fontenc}    % use 8-bit T1 fonts
\usepackage{booktabs}       % professional-quality tables
\usepackage{amsfonts}       % blackboard math symbols
\usepackage{nicefrac}       % compact symbols for 1/2, etc.
\usepackage{microtype}      % microtypography
\usepackage{amsmath,amsfonts,amsthm,commath}
\usepackage{enumerate}
\usepackage{framed}
\usepackage{xspace}
\usepackage{microtype}
\usepackage{nicefrac}
\usepackage[utf8]{inputenc}
\usepackage{CJK}
\usepackage{indentfirst}
\usepackage{booktabs,color} % For formal tables
\usepackage{fancyhdr}
\usepackage{graphicx}
\usepackage[tight,footnotesize]{subfigure}
\usepackage{multirow}
\usepackage{listings}
\usepackage{color}
\usepackage{caption}
\usepackage{algorithm}
\usepackage{algorithmicx}
\usepackage[noend]{algpseudocode}
\usepackage{xspace}
\usepackage{url}
\usepackage{breakurl}

\usepackage{flushend}
\usepackage{xspace,color}

\newtheorem{lemma}{Lemma}[section]

\newtheorem{theorem}{Theorem}[section]

\newtheorem{definition}{Definition}[section]

\newcommand{\he}[1]{{\textsf{\textcolor{red}{[From He: #1]}}}}

\newcommand{\para}[1]{\xspace \smallskip \noindent\textbf{#1}\xspace}
\newcommand{\bx}{\mathbf{x}}
\newcommand{\sysn}{\text{LOCB}\xspace }
\newcommand{\hatt}{\hat{\boldsymbol\theta}}
\newcommand{\ntheta}{\boldsymbol\theta}

\newcommand{\bm}[1]{B_{\ntheta, #1}}

\newcommand{\bthe}{B_{\boldsymbol\theta, i }(m_{i,t}, \delta')}
\newcommand{\bthes}{B_{\boldsymbol\theta, s }(m_{s,t}, \delta')}

\newcommand{\nst}{\mathcal{N}_{s,t}}
\newcommand{\nstt}{\mathcal{N}_{s,t-1}}

\newcommand{\cbri}{CB_{r, i}}

\newcommand{\xt}{\mathbf{x}_t}

\newcommand{\tnst}{\ntheta_{\mathcal{N}_{s,t} }}
\newcommand{\htnst}{\hat{\ntheta}_{\mathcal{N}_{s,t} }}
\newcommand{\cbnst}{CB_{r, \mathcal{N}_{s,t}}}

\newcommand{\hide}[1]{}

\title{Local Clustering in Contextual Multi-Armed Bandits}
\begin{document}

\author{Yikun Ban}
\affiliation{%
  \institution{University of Illinois at Urbana-Champaign}
  \country{}
  }
\email{yikunb2@illinois.edu}

\author{Jingrui He}
\affiliation{%
  \institution{University of Illinois at Urbana-Champaign}
  \country{}
}
\email{jingrui@illinois.edu  }

\begin{abstract}
We study identifying user clusters in contextual multi-armed bandits (MAB). Contextual MAB is an effective tool for many real applications, such as content recommendation and online advertisement. In practice, user dependency plays an essential role in the user's actions, and thus the rewards. Clustering similar users can improve the quality of reward estimation, which in turn leads to more effective content recommendation and targeted advertising. Different from traditional clustering settings, we cluster users based on the unknown bandit parameters, which will be estimated incrementally. In particular, we define the problem of cluster detection in contextual MAB, and propose a bandit algorithm, LOCB, embedded with local clustering procedure.
And, we provide theoretical analysis about LOCB in terms of the correctness and efficiency of clustering and its regret bound.  Finally, we evaluate the proposed algorithm from various aspects, which outperforms state-of-the-art baselines.

\end{abstract}

\maketitle

%\vspace{-1em}
\section{Introduction}
%\vspace{-0.5em}

The recommender system is ubiquitous in online applications. However, in the cold-start setting and the rapid change of recommendation contents,  the conventional approaches that demand sufficient historical records, e.g., collaborative filtering \cite{sarwar2001item, o1999clustering}, usually suffer from the sub-optimal performance~\cite{2010contextual, 2014onlinecluster}. This dilemma between the exploration of new information and the exploitation of empirical feedback also exists in clinical trials \cite{2018durandcontextual, 2020bastanionline}, crowdsourcing \cite{zhou2020crowd, zhou2018unlearn}. Multi-Armed Bandit (MAB) has been extensively studied for online decision making and provides principled solutions for the dilemma of exploration and exploitation ~\cite{auer2002finite,2011improved,bubeck2012regret}.

One of MAB's common applications is the personalized recommendation \cite{2010contextual, 2011improved, 2019improved, 2014onlinecluster,2016collaborative, chu2011contextual, djolonga2013high, tang2015personalized}, such as the recommendation of movies, music, and articles for a user. In contextual MAB, in each round, a set of context vectors is presented to incorporate the side information of recommended items, and an unknown bandit parameter is held for each user to formulate his/her preference (i.e., how the user interacts with the environment). Then, the learner uses some strategy to choose a context vector and receives the corresponding reward. 
In this paper, we consider one common setting that the received reward is computed by a linear function of a context vector and the bandit parameter. \cite{2010contextual, 2011improved,chu2011contextual,dimakopoulou2019balanced,2019improved, 2014onlinecluster,2016collaborative}.

Standard bandit algorithms view each user as an individual and make recommendations only based on the user's own historical rewards, not taking other users' feedback into account ~\cite{2010contextual,2011improved,chu2011contextual,tang2015personalized,djolonga2013high}.
\hide{\he{Please clarify what you mean by collaborative information.}}
In practice, mutual influence among people does exist and plays an essential role in a user's action. Hence, leveraging user dependency is able to improve the quality of recommendation. 
For example, in a music recommendation platform, 
some users are facing a fixed set of songs that is formulated as a set of arms. 
The users who have similar tastes can be clustered into a group; then, when a learner is about to recommend a song (arm) to a user, the songs that have been rated highly by other users in the same group should be taken into account. 
Such applications can be easily found in movie/article/news recommendations.

A line of works ~\cite{2014onlinecluster, 2016collaborative, gentile2017context, 2019improved} has been proposed to incorporate the user dependency in the contextual MAB framework. 
Consider the scenario that many users are facing a fixed set of arms. In each round, a user is given, and the learner needs to pull an arm for the user, obtaining a reward. As the user's bandit parameter is unknown, these works cluster users based on the empirical estimate of the bandit parameter, using a top-down hierarchical clustering procedure. When trying to select the optimal arm, instead of only using the user's own historical rewards in standard bandits, this line of works chooses the arm based on all the users' historical rewards from the same cluster as the current user, thus improving the quality of recommendation.

\begin{figure}[t] 
    \includegraphics[width=0.7\columnwidth]{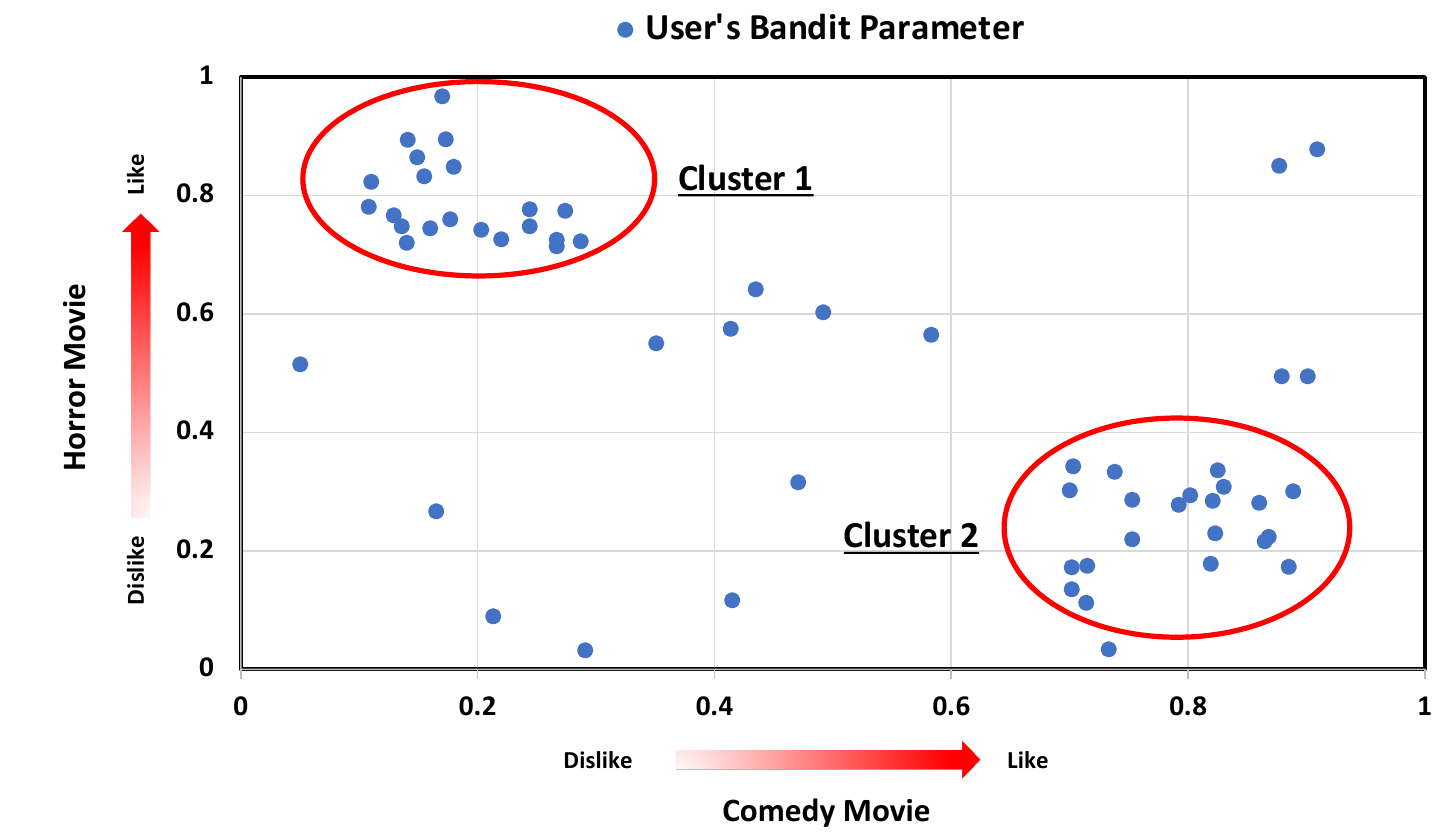}
    \centering
    	\vspace{-1em}
    \caption{ Cluster structure on MovieLens dataset: the users' bandit parameters usually are close to each other rather than being the same. }
       \label{fig:example}
\end{figure}

\begin{figure}[t] 
	\vspace{-1em}
    \includegraphics[width=0.65\columnwidth]{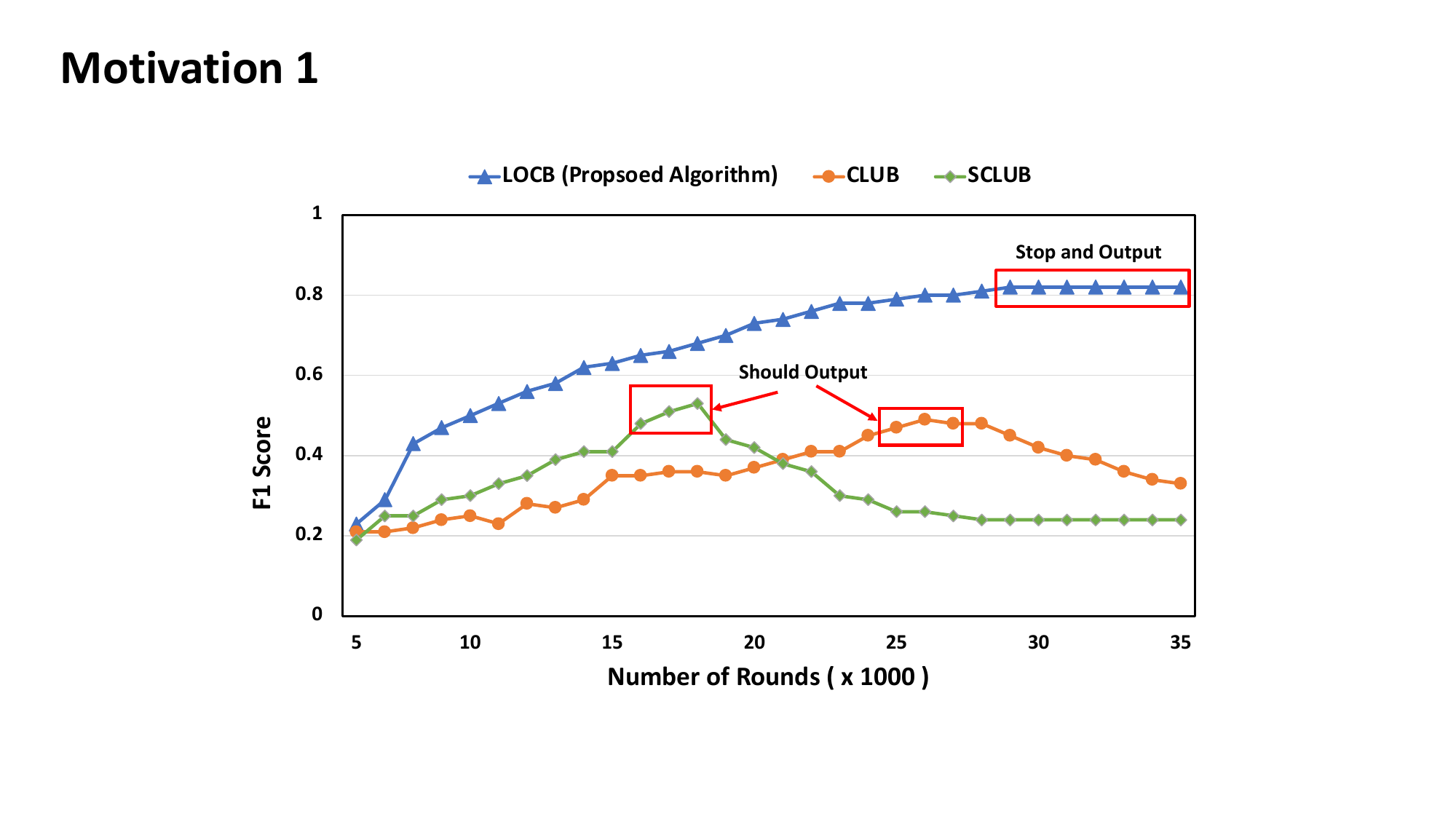}
    \centering
    	\vspace{-1em}
    \caption{ The varying of clustering accuracy for \sysn compared to CLUB and SCLUB on MovieLens dataset.}
       \label{fig:m2}
\end{figure}

However, this line of works suffers from two major drawbacks.
First, they apply a strict assumption to the definition of a cluster ~\cite{2014onlinecluster, 2016collaborative, gentile2017context, 2019improved}, i.e., they consider the users within the same cluster to share exactly the same bandit parameter. In real applications, this assumption is often violated: users may have similar tastes or preferences (represented by bandit parameters), but they hardly have exactly the same ones. For example, Figure \ref{fig:example} shows the two-dimensional mapping of 60 users' bandit parameters on MovieLens dataset, and none of them have the same bandit parameters. Second, existing works~\cite{2014onlinecluster, 2016collaborative, gentile2017context, 2019improved} do not evaluate the quality of the obtained user clusters, and thus are not able to automatically output good user clusters. They maintain either the connected components or sets to represent the current clusters, and update the clusters according to the empirical estimate of bandit parameter in each round. In the clustering process, all the users start in a single cluster, and then gradually partitioned into multiple clusters until each cluster contains one user or a few users with the same bandit parameter (if it exists).
Figure \ref{fig:m2} shows the varying accuracy of CLUB~\cite{2014onlinecluster} and SCLUB~\cite{2019improved} for identifying user clusters.  Although they achieve peak performance at certain rounds, they lack the ability to identify and output the associated user clusters.

To solve the above challenges, in this paper,  we aim to cluster users with similar tastes or preferences, formulated by a set of users with close bandit parameters. This problem is crucial and applicable to many applications. First, it can improve the quality of item recommendation (e.g., movie, music, and product).  When recommending an item to a user, consider other users' preferences in the cluster to which the user belongs, which has been successfully demonstrated by existing works ~\cite{2014onlinecluster, 2016collaborative, gentile2017context, 2019improved}.
Second, identifying clusters can be used for the user recommendation. 
Connecting two users with similar tastes can help them to discover new items and obtain additional insights in the content curation platforms ~\cite{wang2020user, schall2014follow}, because users can collect existing content and provide insights via comments or reviews. For example, Spotify (or Youtube) allows users to create and share their playlist. A user can follow the recommended users and keep track of their listening activities (songs, albums, and playlist).

To identify user clusters in the contextual MAB framework, we propose a bandit algorithm embedded with a clustering procedure, named \sysn (LOcal Clustering in Bandits). 
It can be described as two modules.
One is the Clustering module to find latent clusters among users. Different from the existing global clustering algorithms  ~\cite{2014onlinecluster, 2016collaborative, gentile2017context, 2019improved}, the Clustering module starts with a number of seeds and recursively refines neighbors for each seed. To return the clusters at appropriate rounds, we introduce a termination criterion. Once the criterion is met, the clustering stops and returns multiple clusters that allow for sharing users. 
The other is the Pulling module for the canonical online decision making, which allows the generic integration of user clusters' information. 
Based on the fact that a user may belong to several clusters, the Pulling module is capable of dealing with overlapping clusters, instead of the hard clustering required by previous works  ~\cite{2014onlinecluster, 2016collaborative, gentile2017context, 2019improved}. 
Therefore, in each round, the Pulling module receives the (overlapping) clusters found by the Clustering module and then utilizes them to select an arm and observe the reward.
The key contributions of this paper can be summarized as follows:

(1) \textbf{Problem Definition}: We introduce a user clustering problem in the contextual MAB, which needs weaker assumptions and is applicable to many real-world scenarios.

(2) \textbf{Algorithm}: We propose a bandit algorithm, \sysn, embedded with a local clustering procedure. Different from global online clustering, it is more scalable where the computational cost of each round is proportional to the number of given seeds instead of the number of users. Furthermore, we first study the overlapping clusters in the contextual MAB.

(3) \textbf{Theoretical Analysis}:
We provide three main theorems. The first is the correctness guarantee with respect to clusters returned by \sysn, in order to solve the clustering problem with high probability. The second is the upper bound of the number of rounds needed for the Clustering module to terminate. This bound is $O(n\log n)$ to shows the efficiency of the Clustering module, where $n$ the number of users. Finally, we provide the regret analysis of \sysn to show a regret bound free of the number of clusters, which enables the learner to adjust the number of seeds with a deterministic bound. 

(4) \textbf{Empirical Performance}: We perform extensive experiments on synthetic and real-world datasets from various aspects to evaluate \sysn, including clustering accuracy, regret comparison, and the effect of parameters. \sysn outperforms the state-of-the-art baselines.

The rest of the paper is organized as follows. After briefly introducing the related work in Section 2, we formally present the problem definition in Section 3. The proposed algorithm is introduced in Section 4, and the following theoretical analysis is presented in Section 5. Finally, we show the experimental results on both synthetic and real-world data sets in Section 6. The appendix is placed at the end to include the proofs.

%\vspace{-1em}
\section{Related Work}
%\vspace{-0.5em}

%In this section, we briefly review the related works on contextual bandits.

\hide{
\he{You may also want to discuss user/cluster detection in recommender systems not based on contextual MAB.}}

The multi-armed bandit first proposed by ~\cite{thompson1933likelihood} provides a principled solution for the exploitation-exploration dilemma, which has been adapted to many personalized applications such as advertisement displaying ~\cite{li2010exploitation,chapelle2011empirical, ban2021ee}, recommender system~\cite{2011improved,ban2021multi,ban2021convolutional}, search engine system ~\cite{radlinski2008learning,yue2009interactively}. The traditional non-contextual bandit was studied for various problem setting, such as best arm identification ~\cite{auer2002finite,2010best}, outlier arm identification~\cite{gentile2017context, ban2020generic}, and Top-K arm problems ~\cite{buccapatnam2013multi,kalyanakrishnan2012pac}. In contrast, the contextual bandit was first introduced at ~\cite{auer2002using} and then extended by ~\cite{2010contextual}, in which the arm is represented by a contextual vector instead of a scalar.

\iffalse
In the recommender system, it has been demonstrated that clustering Web users who have similar interests or are geographically close leads to improvements in the performance of content recommendation \cite{chen2008combinational, sarwar2001item}. 
However, in the cold-start settings of a recommender system, namely, the lack of users' accumulated interactions with items, the traditional approaches, e.g., collaborative filtering \cite{sarwar2001item, o1999clustering}, may suffer from sub-optimal performance~\cite{2010contextual, 2014onlinecluster,2016collaborative}.
On the other hand, contextual MAB algorithms turn out to be effective solvers in such settings \cite{2010contextual, 2011improved, 2019improved, 2014onlinecluster,2016collaborative, chu2011contextual, djolonga2013high, tang2015personalized}. However, standard bandits do not consider the mutual influence among users.  

\fi

The idea of exploring dependency among users in the contextual MAB has been studied by a series of works \cite{2014onlinecluster, 2016collaborative, gentile2017context, 2019improved}. CLUB\cite{2014onlinecluster} first considers clustering users based on bandit parameters. It represents the similarity of users by a graph and keeps refining the groups.
To dig out more dependencies, the follow-up work \cite{2016collaborative} clusters both users and items, and leverages the cluster effects together.
\cite{gentile2017context} introduces a context-aware clustering of bandits algorithm that allows each content item to cluster users into groups, where the users from a group have the same reaction to particular items. 
\cite{li2018online, 2018durandcontextual} study a variant of clustering in bandit algorithms that allow recommending a list of items to a user.
Since the above works all assume the users are drawn from a uniform distribution, \cite{2019improved} presents a framework to deal with users with different frequency.
However, as mentioned before, these works do not specify when to return clusters and have no guarantee about the quality of clusters. These limitations will negatively affect the reward estimation, and thus the performance of the overall system.
\cite{qi2022neural} studies the correlation among arms to improve the model's performance.
\hide{
Besides these works, \cite{wu2016contextual,buccapatnam2013multi, cesa2013gang} take advantage of a network of social relationships to make a better recommendation.   \cite{korda2016distributed} considers the linear bandit problems in peer to peer networks and proposes two distributed algorithms for clustering of confidence ball. However, they assume the social relations is known in advance.}

Cluster detection has been studied for decades, and many algorithms were proposed to solve this problem. We refer the reader to several survey papers \cite{o1999clustering,xu2005survey}. Existing methods include: local spectral algorithms \cite{von2007tutorial, 2012A}, graph-based clustering \cite{graphdiff, yikun2019no}, modularity optimization \cite{newman2004finding},  motif-based clustering  \cite{zhou2021high,fu2020local,zhou2017local}, and so on.  
Different from traditional clustering, we cluster users based on the unknown bandit parameters rather than the known links or attributes.

%\vspace{-1em}
\section{Problem Definition}
%\vspace{-0.5em}

In this section, we formulate the joint problem of user clustering and contextual MAB, where the learner aims to cluster users with similar bandit parameters.

Let $N = \{1, \dots, n \}$ be a set of $n$ users. 
At each round $t = 1, 2, \dots$, $T$, the learner receives a user $i_t \in N$ and observes a set of $k$ context vectors  $\mathbf{X}_t = \{\mathbf{x}_{1,t}, \mathbf{x}_{2,t}, \dots, \mathbf{x}_{k,t}\}$ associated with $k$ arms.
Then, the learner chooses some $\bx_{a, t} \in \mathbf{X}_t $ to recommend to the user $i_t$ and obtains the reward $r_t$.
For each $\bx_{a,t} \in \mathbf{X}_t $, $\bx_{a,t} \in \mathbb{R}^{d}$, \hide{\he{What is $d$?} }it summarizes the side information of arm $a$ at current round $t$.      
Suppose that the user $i_t$ of each round is drawn uniformly from $N$, and s/he is associated with an unknown bandit parameter $\ntheta_{i_t}\in \mathbb{R}^{d}$, reflecting how $i_t$ interacts with the environment.    
In standard linear contextual bandit problems ~\cite{2010contextual,2014onlinecluster,wu2016contextual},  the reward $r_t$ is governed by a noisy version of an unknown linear function of $\bx_{a,t}$ and $\ntheta_{i_t}$:
\[ 
r_t =   \ntheta_{i_t}^{\intercal} \bx_{a,t}  + \eta_t,
\]
where $\eta_t$ is a noise with zero-mean and $\sigma$-bounded variance, drawn from a Gaussian distribution $\mathcal{N}(0, \sigma^2)$.
%\he{What is $\sigma$?}

The users with similar behaviors are considered to form a cluster.
Without any side information of users, we measure their similarity by comparing their associated bandit parameters $\ntheta$ (unknown).
In contrast with using the strict assumption of existing works ~\cite{2014onlinecluster,2016collaborative,gentile2017context, 2019improved} that the users from a cluster share the same $\ntheta$, we allow some deviation in $\ntheta$ of users from the same cluster. More specifically, the deviation is upper bounded by a threshold $\gamma$ predefined by the learner, which is usually a small constant.
Formally, we introduce a generic definition, $\gamma$-Cluster.

\begin{definition} [ $\gamma$-Cluster]
Given a subset of users $\mathcal{N} \subseteq N $ and a threshold $\gamma>0$, $\mathcal{N}$ is considered as a $\gamma$-Cluster if it satisfies 
\[
\forall i, j \in \mathcal{N},    \Arrowvert \mathbf{\ntheta}_i   - \mathbf{\ntheta}_{j}  \Arrowvert  < \gamma.
\]
\end{definition} 

In this paper, the first objective is to design an efficient algorithm to recover the clusters among users, such that the set of clusters returned by the algorithm are true $\gamma$-clusters with probability $1-\delta$, where $\delta$ is a small constant. 

The task of finding $\gamma$-clusters is challenging due to the following reasons.

(1) Whether the return clusters are true $\gamma$-clusters. As the bandit parameter is unknown for each user, the learner needs to use the estimation to cluster after observing rewards in each round. This is a traditional obstacle in the contextual MAB \cite{2014onlinecluster,2011improved}.

(2) When to return the clusters. Playing too many rounds leads to high costs, but playing only a few rounds introduces serious uncertainties. This is a new exploration-exploitation dilemma in this problem.

(3) Overlapping clusters. It is common that multiple clusters are overlapping, where a user may belong to more than one $\gamma$-clusters. The existing works  ~\cite{2014onlinecluster, 2016collaborative, gentile2017context, 2019improved} focus on hard clustering and may fail in the presence of overlapping.

In addition to identifying the user clusters, we also aim to minimize the accumulated regret as the goal in standard bandits. More specifically, after $T$ rounds, the accumulated regret for all users is defined as,
\[
 \mathbf{R}_T =  \mathbb{E} [ \sum_{t =1 }^{T}R_t] = \sum_{t=1}^{T}(\ntheta_{i_t}^{\intercal} \bx_{t}^* -  \ntheta_{i_t}^{\intercal} \bx_{t})
\] 
where $i_t$ is the served user in round $t$,  $\bx_{t}^* = \arg \max_{\bx_{a,t} \in \mathbf{X}_t}  \ntheta_{i_t}^{\intercal} \bx_{a, t}$, and $\bx_{t}$ is the pulled arm in round $t$ in practice.

%(4) The number of clusters in unknown. This further the change

%\vspace{-0.5em}
\section{\sysn: Local Clustering in bandits}\label{sec:3}
%\vspace{-0.5em}

In this section, we introduce the proposed algorithm, \sysn, for detecting and exploiting underlying clusters among users in the contextual MAB.
It has two cooperative modules: the Clustering and Pulling.
In the round $t-1$, after pulling an arm determined by the Pulling module and observing the reward $r_{t-1}$, the Clustering module updates the membership for each cluster based on $r_{t-1}$. Then, in the round $t$, given the clusters provided by the Clustering module that allows for overlapping, the Pulling module finds the optimal cluster for the serving user $i_t$ and selects an arm, obtaining the reward $r_t$.  Next, we first elaborate on the Clustering module and then the Pulling module.

As the standard contextual bandit~\cite{2010contextual,wu2016contextual}, in each round $t$, \sysn needs to compute the estimation $\hat{\ntheta}_{i_{t},t}$ of $\ntheta_{i_t}$ for $i_t \in N$ in round $t$, after pulling an arm $\bx_{t}$ and observing the reward $r_t$:
\begin{equation}
\hatt_{i_t,t} = {\mathbf{A}_{i_t,t}}^{\mathbf{-1}}\mathbf{b}_{i_t,t}, \ \ 
{\mathbf{A}_{i_t,t}} = \mathbf{I} + \sum_{\bx_{i_t}  \in \mathcal{H}_{i_t,t}} \mathbf{x}_{i_t}\mathbf{x}_{i_t}^{\intercal}, \ \ 
\mathbf{b}_{i_t,t} = \sum_{ (\bx_{i_t}, r_{i_t}) \in \mathcal{H}_{i_t,t}  } \mathbf{x}_{i_t}r_{i_t},
\end{equation}
where $\mathbf{I}$ is a $d \times d$ identity matrix and $\mathcal{H}_{i_t,t}$ represents the historical data of user $i_t$ up to round $t$.

\subsection{Clustering Module}

This module is a seed-based clustering algorithm. It randomly chooses a set of seed users, denoted by $S = \{s_1, \dots, s_K \}, S \subseteq N $,  and progressively learns the neighbors of each seed user. Supposing in the round $\hat{T}$, it terminates and then outputs $|S| = K$ clusters, denoted by $\mathbf{N}_S = \{ \mathcal{N}_{s_1,\hat{T}}, \dots, \mathcal{N}_{s_{K},\hat{T}} \}$, for each $\mathcal{N}_{s,\hat{T}} \in \mathbf{N}_S,  \mathcal{N}_{s,\hat{T}} \subseteq N$.

The neighborhood between two users should be determined when it is confident about whether their bandit parameters are close enough. We achieve this goal by using the confidence interval of $\hatt_{i,t}$ for each user $i \in N$. If the significance levels of these intervals are carefully set, we can safely terminate the module with a certain criterion while guaranteeing that the cluster returned for each seed is a true $\gamma$-cluster with probability at least $1-\delta$. 

The general definition of confidence interval for $\hatt_{i,t}$ is defined as:
\[
\mathbb{P}\left( \forall t\in [T],  \Arrowvert \hatt_{i,t} -  \boldsymbol\theta_i \Arrowvert >  \bthe \right) < \delta',
\]
where  $[T] = \{1, 2, \dots, T\}$,  $\bthe$ is an upper confidence bound and $m_{i,t}$ is the number of times that $i$ has been served up to $t$.
With probability $1-\delta$, to ensure $\forall t, \forall i \in N, \hat{\ntheta}_{i,t}$ is within the confidence interval, the significance level $\delta'$ should be set as $\delta/n$ (Lemma \ref{lemma1}).

If two users belong to the same $\gamma$-cluster, we call them neighbors. 
Thus, given a user $i \in N$ and a seed user $s \in S$, we consider $i$ as $s$'s potential neighbor if their confidence intervals are overlapping, which is formally defined as:
\begin{equation}\label{eq:1}
\Arrowvert \hatt_{i, t} - \hatt_{s, t} \Arrowvert \leq B_{\boldsymbol\theta, i }(m_{i,t}, \delta')  + B_{\boldsymbol\theta, s}(m_{s,t}, \delta').  
\end{equation}

Cluster module updates and keeps the potential neighbors of each seed in each round, until it is confident that the current potential neighbors are real neighbors. As $\gamma$ is predefined by the learner, the number of rounds for exploring clusters varies. Let $\mathcal{N}_{s,t}\subseteq N$ denote the set consisting of the seed user $s$ and $s$'s potential neighbors. We provide the termination status when the learner is confident that $\mathcal{N}_{s,t}$ is a $\gamma$-cluster and the Clustering module should stop exploring $\mathcal{N}_{s,t}$. We define the termination status as: return $\mathcal{N}_{s,t}$ if
\begin{equation}\label{eq:2}
 \sup \{\bthe : i \in \mathcal{N}_{s,t}\} < \frac{\gamma}{8} \cdot \tau, 
\end{equation}
where $\tau$ is a tuning parameter with respect to theoretical criterion in Theorem \ref{theo:1}.

Algorithm \ref{alg:main} Lines 25-34 describe the high-level idea of the Clustering module. 
Give a set of seeds $S$ and $\gamma$ to \sysn, where $S$ is randomly chosen from $N$ and the number of seeds $|S|$ will be discussed at end of this section. In the initialization, for each $s \in S$, we set $\mathcal{N}_{s,t} = N$ when $t = 0$ (Lines 3-4), because we consider all the users as the seed user's potential neighbors before we receive any information about them. In each round $t$, after pulling an arm determined by Pulling module and observing the reward,  $\hat{\ntheta}_{i_t,t}$ of the served user $i_t$ is computed (Line 26). Then, we compare  $\hat{\ntheta}_{i_t,t}$ with $\hat{\ntheta}_{s,t}$ for each $s \in S$ to determine whether $i_t$ is $s$'s potential neighbor (Lines 28-29). If not, we remove $i_t$ from $\mathcal{N}_{s,t}$ (Line 30). Therefore, the potential neighbors of each seed is updated in each round. Cluster module will stop exploring $s$ if $\mathcal{N}_{s,t}$ meets the stop criterion (Eq.(\ref{eq:2})) and we remove $s$ from $S$ (Lines 33-34). When $S$ is an empty set, the Clustering module terminates and outputs the set of clusters $\{\mathcal{N}_{s,t}: s \in S \}$ (Lines 20-23).

\begin{algorithm}[th]
\renewcommand{\algorithmicrequire}{\textbf{Input:}}
\renewcommand{\algorithmicensure}{\textbf{Output:}}
\caption{ \sysn }\label{alg:main}
\begin{algorithmic}[1]
\Require $\gamma$, a set of seeds $S$, exploration parameters $\alpha, \tau$ 
\Ensure Clusters $\mathbf{N}_S$
\For{ each $i \in N$}
\State $\mathbf{A}_{i,0} \leftarrow \mathbf{I}$, $\mathbf{b}_{i,0} \leftarrow \mathbf{0}$, $m_{i,0} \leftarrow 0$
\EndFor
\For{ each $s \in S$}
\State $\mathcal{N}_{s,0} \leftarrow N$
\EndFor

\For{$t \leftarrow 1, 2, \dots$}
\State receive $i_t \in N $ and obtain $\mathbf{X}_t \leftarrow \{\mathbf{x}_{1,t}, \mathbf{x}_{2,t} \dots, \mathbf{x}_{k,t}\}$
\For{ each $s \in S$}
\State $S_t(i_t) \leftarrow \emptyset$
\If {$ i_t  \in \mathcal{N}_{s,t-1}$}:
\State $S_t(i_t) \leftarrow  S_t(i_t) + \{ s \}$
\EndIf
\EndFor
\For {each $s \in S_t(i) $}
\State $\htnst \leftarrow \frac{1}{|\nstt|} \sum_{j \in \nstt} \mathbf{A}_{j,t-1}^{-1} \mathbf{b}_{j,t-1}$
\EndFor
\State $\bx_t  \leftarrow   \arg \max_{\bx_{a,t} \in \mathbf{X}_t} \max_{s \in S(i_t)}  \left( \htnst^{\intercal} \bx_{a,t} + \cbnst \right) $, where  $  \cbnst  \leftarrow \frac{1}{|\nstt|} \sum_{j \in \nstt} \alpha \sqrt{\bx_{a,t}^{\intercal} \mathbf{A}_{j,t-1}^{-1}   \bx_{a,t}}$
\State  pull $\bx_t$ and observe reward $r_t$
\State $\mathbf{A}_{i_t,t} \leftarrow \mathbf{A}_{i_t, t-1} +\mathbf{x}_t \mathbf{x}_t^{\mathbf{-1}},  \ m_{i_t, t} \leftarrow m_{i_t, t-1} + 1 $
\State $\mathbf{b}_{i_t,t} \leftarrow \mathbf{b}_{i_t,t-1} + r_t \mathbf{x}_t$
%\State $m_{i,t} \leftarrow m_{i,t}+1$
\For {each $ i \in N  \wedge  i \not = i_t$}  
\State $ \mathbf{A}_{i,t} \leftarrow \mathbf{A}_{i, t-1}$
\State $\mathbf{b}_{i,t} \leftarrow \mathbf{b}_{i, t-1}$
\EndFor
\If{ $|S| > 0$ }
\State \textsc{Clustering}$(i_t, S)$
\Else 
\State \textbf{Output} $\mathbf{N}_S \leftarrow \{\mathcal{N}_{s,t}: s \in S \}$ \# \ Clustering module terminates 
\EndIf
\EndFor

\\

\Procedure{Clustering}{$i_t, S$}

\State $\hatt_{i_t,t} \leftarrow {\mathbf{A}_{i_t,t}}^{\mathbf{-1}} \mathbf{b}_{i_t,t}$
\For{each $s \in  S$}
\State $\hatt_{s,t} \leftarrow {\mathbf{A}_{s, t}}^{\mathbf{-1}} \mathbf{b}_{s, t}$
\If {$ \Arrowvert \hatt_{i_t, t} - \hatt_{s, t} \Arrowvert > B_{\boldsymbol\theta, i_t }(m_{i_t,t}, \delta')  + B_{\boldsymbol\theta, s}(m_{s,t}, \delta')   $}
\State  $\mathcal{N}_{s,t} \leftarrow  \mathcal{N}_{s,t-1} - \{i_t\}$ \#  \  Remove $i_t$ from $s$'s potential neighbors
\Else
\State  $\mathcal{N}_{s,t} \leftarrow  \mathcal{N}_{s,t-1} \cup  \{i_t\}$ \#  \ To ensure $i_t$ is in  $\mathcal{N}_{s,t} $
\EndIf
\If{ $\sup \{\bthe : i \in \mathcal{N}_{s,t}\} < \frac{\gamma}{8} \cdot \tau$  }  
\State $S \leftarrow S - \{s\}$  \ \# \ $\mathcal{N}_{s,t}$ is ready to return and remove $s$ from $S$
\EndIf
\EndFor
\EndProcedure
\end{algorithmic}
\end{algorithm}

%where $t \% n =0$ guarantees that each user is observed with the same number of rounds.

\subsection{Pulling Module}

In the last sub-section, we present how the Clustering module finds the cluster for each seed. Given a set of seeds $S$, Clustering module holds a set of clusters in each round $t$, denoted by $\mathbf{N}_{S,t} = \{\mathcal{N}_{s,t}: s \in S \}$. In this section, we will present how the Pulling module utilizes $\mathbf{N}_{S,t}$ in the decision making of the contextual MAB. 

First, as the standard UCB-based (upper confidence bound) bandit \cite{2010contextual,2011improved,2014onlinecluster,gentile2017context}, given an arm $\bx_{a,t}$, we need to define a confidence interval for the estimated reward $\hatt_{i,t}^{\intercal} \bx_{a,t}$ of $\ntheta_{i}^{\intercal} \bx_{a,t}$ for each $i \in N$. The confidence interval is defined as:
\[
\mathbb{P}\left( \forall t \in [T],   |  \hatt_{i,t}^{\intercal} \bx_{a,t}  -  \ntheta_{i}^{\intercal} \bx_{a,t} | >  CB_{r, i}\right) < \delta',
\]
where  $CB_{r, i} = \alpha  \sqrt{ \bx_{a,t}^{\intercal} \mathbf{A}_{i,t-1}^{-1}   \bx_{a,t}  }$ and $\alpha$ is a suitable function satisfying $O(\sqrt{d \log t})$~\cite{gentile2017context, 2011improved, 2010contextual}.

Consider a current cluster $\mathcal{N}_{s,t}$ (represented by $\mathcal{N}_{s,t-1}$ in Algorithm \ref{alg:main}). We define a bandit parameter $\tnst$ for  $\mathcal{N}_{s,t}$ to represent the integration of the included users' bandit parameters, formally:
\begin{equation}
\tnst = \frac{1}{|\mathcal{N}_{s,t}|} \sum_{i \in \mathcal{N}_{s,t}} \ntheta_{i}.
\end{equation}
As $\tnst$ is also unknown, we compute its estimation as:
\begin{equation}
\htnst = \frac{1}{|\mathcal{N}_{s,t}|} \sum_{i \in \mathcal{N}_{s,t}} \hatt_{i,t}.
\end{equation}

Based on Lemma \ref{lemma4} in Appendix, we have the confidence interval for $\htnst^{\intercal} \bx_{a, t}$, formally defined as:
\[
\mathbb{P}\left( \forall t \in [T],  |  \htnst^{\intercal} \bx_{a,t}  -  \tnst^{\intercal} \bx_{a,t} | >  \cbnst \right) < \delta',
\]
where $\cbnst = \frac{1}{|\mathcal{N}_{s,t}|} \sum_{i \in \mathcal{N}_{s,t}}CB_{r, i}$.

Similar to the existing works ~\cite{2014onlinecluster, 2016collaborative, gentile2017context, 2019improved}, given a served user $i_t$ and one of $i_t$'s clusters $\nst$, we determine the arm by $\htnst$ rather than $\hatt_{i_t,t}$, using the following criteria:
\[
\bx_{t} = \arg \max_{\bx_{a,t} \in \mathbf{X}_t} \htnst^{\intercal} \bx_{a,t} + \cbnst.
\]
However, a user may belong to more than one clusters. In particular, when $|S|$ is large, it is highly likely that a user belongs to multiple clusters returned by the Clustering module. To find the optimal cluster from these candidate clusters, we propose the following criterion
to find the cluster with maximal potential. Let $ S_t(i_t)$ represent the clusters that $i_t$ belongs to in round $t$ \hide{\he{$S(i)$ should be $S_t(i)$ instead, right?}}, $ S_t(i_t) = \{ s: s \in S \wedge i_t \in \mathcal{N}_{s,t}    \}$.
Then, the Pulling module selects an arm by:
\begin{equation}\label{eq:over}
\bx_{t} = \arg  \max_{\bx_{a,t} \in \mathbf{X}_t} \max_{s \in S_t(i_t)} \left( \htnst^{\intercal} \bx_{a,t} + \cbnst \right).
\end{equation}
\hide{\he{This equation is problematic.}}
Note that if the user $i_t$ does not belong to any cluster in round $t$, i.e., $S_t(i_t) = \emptyset$, then the selection criterion follows the standard UCB-based bandit \cite{2010contextual}, determined by $  \arg  \max_{\bx_{a,t} \in \mathbf{X}_t} (\hatt_{i_t,t}^{\intercal}\bx_{a,t} +  CB_{r, i})$. This is not shown in Algorithm \ref{alg:main} because of limited space.

Algorithm \ref{alg:main} Lines (1-19) describe the workflow of the Pulling module.
Lines 1-4 show the initialization for each user and each seed user.  In each round, after observing the served user $i_t$ and context vectors $\mathbf{X}_t$, it first finds the clusters $i_s$ belongs to among the clusters kept by the Clustering module, represented by $S_t(i_t)$ (Lines 5-10). Then, we decide the arm to pull by the criterion Eq.(\ref{eq:over}) (Lines 11-13). With the observed reward $r_t$, we update the parameters for each user (Lines 15-19). At the end of each round, the Clustering module starts to update the clusters for each seed (Lines 20-23).

\para{Selection of Seeds}.
Similar to traditional seed-based clustering algorithms ~\cite{kanungo2002efficient, yin2017local, fu2020local}, the number of seeds affects the performance of \sysn. In general, the more seeds the algorithm is given, the higher chance it has to find good clusters. Therefore, it is encouraged for the learner to have a relatively large number of seeds. In many cases, we can simply set $S = N$ to achieve the best performance the Clustering module may reach.
In this case, each user will be treated as a center, and the Module module explores each user's potential neighbors in each round. 
For the isolated user, one will hold a cluster that only includes him/herself.
For those users who are close to each other, they will hold a $\gamma$-cluster respectively in the end, while these clusters are very likely to be overlapping.
Therefore, the Pulling module is designed to deal with overlapping clusters. It can find the optimal cluster among candidate clusters for a user.  Therefore,  the cooperation of the Pulling module with the Clustering module can effectively alleviates the challenge of finding good seeds.

\hide{\he{If all the users are used as seeds, will it happen that each user will belong to its own cluster? You need to provide some discussions here, as the reviewers are likely to ask this question based on your description.}}

\section{THEORETICAL ANALYSIS}
In this section, first, we introduce the two theorems to show \sysn's effectiveness and efficiency for solving $\gamma$-cluster detection problem. Then, we provide the detailed regret analysis of \sysn.

We first provide the theoretical analysis with respect to the detected clusters by \sysn as follows.

\begin{theorem}[Correctness] \label{theo:1}
Given a threshold $\gamma$ and  a set of seeds $S \subseteq N$, for each $s \in  S$, let $\mathcal{N}_{s}$ represent the cluster output by \sysn with respect to $s$.  The terminate criterion of Clustering module is defined as:
\[ \sup \{\bthe : i \in \mathcal{N}_{s,t}\} < \frac{\gamma}{8}.\]
Then, with probability at least $1-\delta$, after the Clustering module terminates, for each $s \in S$, it has
\[
\forall i, j \in \mathcal{N}_s,  \Arrowvert \ntheta_i   - \ntheta_{j}  \Arrowvert  < \gamma. 
\]
\end{theorem}

The details of the proof are provided in Appendix. Before proving Theorem \ref{theo:1}, we need to set the confidence interval properly for each user in each round, in order to make sure that the estimations are within the confidence interval during the clustering procedure, as shown in Lemma \ref{lemma1}.

\begin{lemma} \label{lemma1}
At round $t = 1, 2, \dots, T$, suppose that for any user $i \in N$, we have an upper confidence bound $\bthe$ that satisfies:
\[
\mathbb{P}( \forall t \in [T],  \Arrowvert \hatt_{i, t} - \ntheta_{i} \Arrowvert > \bthe) < \delta',
\]
where  $\delta'$ is a varying confidence level with respect to $\delta$ and $t$.
Define the random event 
\[
\mathcal{E} =  \left \{  \bigwedge_{\forall t\in [T], \forall i \in N}  \Arrowvert \hatt_{i, t} - \ntheta_{i} \Arrowvert \leq \bthe \right \}.
\]
Let  $I_T = [i_1, i_2, \dots, i_T]$ be the sequence of served users in each round up to $T$.  If $\delta'$ is defined as
\[
\delta' = \frac{ \delta}{n},
\]
then, the probability of $\mathcal{E}$ happening is higher than $1-\delta$,
\[ 
i.e., \mathbb{P}(\mathcal{E} | I_T) \geq  1- \delta.
\]
\end{lemma}

Second, we introduce an upper bound on the number of rounds needed for the Clustering module to terminate. 
First, we need to define $\bthe$, following the upper confidence bound in ~\cite{2014onlinecluster}.

\begin{lemma}~\cite{2014onlinecluster} \label{lemma:ucb}
For each round $t$, let the context vectors $\mathbf{X}_t = \{\bx_1, \cdots, \bx_{k} \}$ be generated i.i.d (conditioned on $i_t, k$ and past data $\{i_t', \mathbf{X}_{t'}, r_{t'} \}_{t' = 1}^{t}$) from a random vector $\mathbf{X}$ such that $|\mathbf{X} || =1$ and $\mathbb{E}[\mathbf{XX}^{\intercal}]$ is full rank with minimal eigenvalue $\lambda > 0$. 
Then, given a user $i \in N$,  with probability $1-\delta$, for any $t \in [T]$, it has    
\[
\Arrowvert \hatt_{i, t} - \ntheta_{i} \Arrowvert \leq \bthe
\] 
where 
\[
\bthe = 
\frac{ \sigma \sqrt{2d \log t + 2\log(2/\delta')}+1}{\sqrt{1+ h(m_{i,t}, H)} },
\] 
\[
h(m_{i,t}, H) = \left ( \frac{\lambda m_{i,t}}{4} - 8\log(\frac{ m_{i,t}+3}{H}) - 2 \sqrt{ m_{i,t} \log \left ( \frac{ m_{i,t} + 3}{ H} \right ) } \right ),
\]
and $H =\delta' / 2nd$. 
\end{lemma}

With this UCB, we have the following theorem.

\begin{theorem} \label{theo:2}
Suppose each user is evenly served and $m_{i,t} \geq \frac{2 \times 32^2 }{\lambda^2} \log \left ( \frac{2nd}{\delta'} \right) \log \left ( \frac{32^2}{\lambda^2} \log\left( \frac{2nd}{\delta'} \right) \right)$ for any $i \in N$. Then, with probability at least $1 - \delta$, the  number of rounds $\hat{T}$ needed for the Clustering module to terminate is upper bounded by
\[
\hat{T} < \frac{2nd}{C} \log \frac{nd}{C}  +  \frac{2n}{C}\left(  \log(\frac{2^{(d+1)}n}{\delta}) - \frac{\gamma^2-256}{512\sigma^2} \right)  + n.
\] 
where $
C = \frac{ \lambda \gamma^2 }{16^3  \sigma^2 }.$

\end{theorem}

The above theorem provides the upper bound of the cost for \sysn to output $\gamma$-clusters.
Denote this upper bound by $\bar{U}$. In practice, $n$ usually is a large number, then $\bar{U}$ becomes $O (n \log n)$. If $d$ is also a large number, $\bar{U}$ becomes $O (nd \log nd)$. 

Finally, we provide the regret bound for \sysn.

\begin{theorem}\label{tho:bound}
Suppose that each user is evenly served. Given $\gamma$ and a set of seeds $S$, after $T > \hat{T} $ rounds, the accumulated regret of \sysn can be upper bounded as follows:
\[
\begin{aligned}
\mathbf{R}_T &\leq  \left[ \sqrt{nT}  \cdot
\sqrt {2d\log (1+T/dn) } \cdot O \left( \sqrt{ d  \log  \left(T/\delta  \right ) }  \right)\right] \\
& + \left(T - O\left(nd \log nd \right)\right) \gamma +  O\left(nd \log nd \right) \cdot O \left( \sqrt{ d  \log  \left(Tn/\delta  \right ) }  \right).
\end{aligned}
\]
\end{theorem}

This upper bound is composed of two terms. The first term is the usual $\sqrt{T}$-style term in linear bandit regret analysis ~\cite{auer2002using,chu2011contextual,2011improved}. Note that this bound does not depend on the number of seed $|S|$, as the clusters are allowed to be overlapping. This indicates that even though \sysn is given with plenty of seeds (e.g., $S = N$), the regret bound will not decay \hide{\he{This sentence is confusing.}}. The second term is the accumulated regret caused by the deviation between the cluster center and served user. This bound happens when $ \|\tnst - \ntheta_{i} \| = \gamma, \forall i \in N$. However, this is too pessimistic. In practice, there usually exist some $\gamma$-clusters for a user $i$ where $\tnst \approx \ntheta_{i}$. If $ \forall t,   \forall i \in N, \|\tnst - \ntheta_{i} \| \rightarrow 0$, the second term will decrease close to zero.
\hide{\he{Will the second term be exactly 0 if you use all the users as seeds?}}

The key difference of the bound in Theorem \ref{tho:bound} from the regret analysis of existing works ~\cite{2014onlinecluster, 2016collaborative, gentile2017context, 2019improved} is the assumption of clusters.  The regret bounds of previous works rely on the number of non-overlapping clusters denoted by $m$ in which each user shares the same bandit parameter. This assumption corresponds to $m$ linear bandits if $m$ is known. However, Unlike this strong assumption, we allow the clusters to overlap, and thus our bound still depends on the number of users. 
%Moreover, instead of using the estimated user's bandit parameter, this bound utilizes the estimation of the cluster center, to be more applicable in practice. 

%\vspace{-1em}

\section{Experiments}
%\vspace{-1em}
To evaluate \sysn from various aspects, we divide experiments into three parts to evaluate its clustering accuracy, accumulated regret, and the effect of input parameters. We compare $\sysn$ with standard bandit algorithms as well as online clustering of bandit algorithms on one synthetic and three real-world datasets.
First, we briefly introduce the four datasets used in experiments.

(1) \textbf{Synthetic.}
We synthesize a dataset with $100$ users and $5$ clusters, where the size of each cluster is randomly chosen from [5 : 40].  Each user $i$ has a separate parameter $\ntheta_i$ and the users within a cluster $\mathcal{N}$ satisfy $\forall i, j \in \mathcal{N}, \Arrowvert \ntheta_i  - \ntheta_j  \Arrowvert < 0.2$. Both $\ntheta_i$ and context vector $\bx_{a,t}$ have $d = 5$ dimensions drawn from a standard Gaussian distribution. Then, they are appended with one more dimension with constant 1, and transformed by $ x \leftarrow \left( \frac{x}{\sqrt{2}\Arrowvert x \Arrowvert}, \frac{1}{\sqrt{2}} \right)$, to guarantee that $\langle \ntheta_i, \bx_{a,t}\rangle$ lies in $[0,1]$.

(2) \textbf{Yelp}\footnote{https://www.yelp.com/dataset} is a dataset released in Yelp dataset challenge. It contains 4.7 million rating records from 1.18 million users to  1.57 $\times 10^5$ restaurants. Each restaurant is represented by a feature embedding vector $\bx_{a,t} \in \mathbb{R}^{10}$ with respect to its attributes and categories. 
We generate the reward by using the restaurant's gained stars scored by the users. In each review record, if the user scores the restaurant more than 3 stars (5 stars totally), the reward $r_t = 1$; Otherwise, the $r_t = 0$. We set the user pool $|N| = 100$ by selecting the top 100 users with the most reviews. And we set the arm pool $|\mathbf{X}_t|= 10$ as follows: given a user $i$, we pick one restaurant with non-zero reward according to the whole records in the dataset, and then randomly pick the other $9$ restaurants with zero rewards.

%Same there are five clusters among users.
 
(3) \textbf{MovieLens} \cite{harper2015movielens} is a dataset consisting of $25$ million ratings of $6 \times 10^4$ movies from $1.6 \times 10^5$ users. Each movie is represented by an embedding vector $\bx_{a,t} \in \mathbb{R}^{10}$ with regard to the genres and historical records. Similarly, a rating event is represented by $\{\bx_{a,t}, i, r_t \}$, where the reward $r_t$ is $1$ if the movie $\bx_{a,t}$ obtains more than 3 stars from the user $i$ (Otherwise $r_t = 0$). We set user pool $|N| = 100$ and the arm pool $|\mathbf{X}_t|= 10$.

(4) \textbf{Yahoo} ~\cite{2010contextual} is a recommendation dataset containing $45$ million user visits to `Yahho! Today Module' across ten days.
Each visit includes one user and ten candidate articles in which both user and article are represented by a six-dimension feature vector. Here, we choose the contextual vector $\bx_{a,t}$ of the first article in each visit as the arm. If the user clicks the this article, the reward $r_t$ is $1$; Otherwise, $r_t$ is $0$.
However, this dataset does not provide user identities. Therefore, following the previous works \cite{2016collaborative, gentile2017context}, we use the $K$-means to cluster user vectors where each cluster is thought of as a user. In the beginning, we set $K=500$ to determine user identities and select the top $100$ users with the most visit records to be the user pool $|N| = 100$. Then, we set the arm pool $|\mathbf{X}_t| =10$ following the above selection strategy.

Across the all experiments, we set $\delta = 0.1$, run each experiments $5$ times, and report the average results.

%\vspace{-1em}
\subsection{Accuracy of Clustering}
%\vspace{-0.5em}

In this sub-section, we evaluate the accuracy of \sysn for the user clustering.
Since the problem setting is new, there are no existing methods focusing on $\gamma$-cluster detection in the contextual MAB. Therefore, we design four baselines for comparison.

\begin{enumerate}
\item 
Naive CLUB (\textbf{N-CLUB}). CLUB\cite{2014onlinecluster} regards connected components as user groups and refines groups gradually. However, it cannot determine when the group is good enough. Therefore, we terminate it when the found groups have not changed in the last consecutive $10/ \delta$ rounds. This baseline reflects how well CLUB can solve the problem with a heuristic termination condition.

\item
Same Termination CLUB (\textbf{ST-CLUB}). We terminate CLUB when \sysn stops, i.e., ST-CLUB and \sysn have exactly the same number of rounds. This baseline reflects the accuracy of \sysn compared to CLUB with the same termination condition.

\item
Same Termination SCLUB (\textbf{ST-SCLUB}). Similarly, we terminate SCLUB \cite{2019improved}  when \sysn stops. This baseline reflects the accuracy of \sysn compared to SCLUB with the same termination condition.

\item
Naive \sysn (\textbf{N-\sysn}). We terminate \sysn when the size of detected group has not changed in the last consecutive $10/ \delta$ rounds. This baseline reflects how well the problem can be solved by the proposed method with a heuristic termination condition.
\end{enumerate}

\para{Obtain ground-truth clusters.}
As the three real-world datasets do not provide the bandit parameter vector for each user, we need to calculate the expectation of the bandit parameter according to each dataset. Assume in a dataset,  a user $i$ totally has $T_i$ records represented by $\{\bx_{t}, i ,r_t\}|_{t=1}^{T_i}$. Then, we compute the expectation $\ntheta_i$ for $i$ by using the standard estimation in contextual MAB, as follows:
\[
\ntheta_{i} = {\mathbf{A}_{i,T_i}}^{\mathbf{-1}}\mathbf{b}_{i,T_i}, \ \ 
{\mathbf{A}_{i,T_i}} = \mathbf{I} + \sum_{t =1}^{T_i} \mathbf{x}_{t}\mathbf{x}_{t}^{\intercal}, \ \ 
\mathbf{b}_{i,T_i} = \sum_{t = 1}^{T_i} \mathbf{x}_{t}r_{t}.
\]
Here, we calculate all the available records for each user of each dataset. With these known bandit parameters, we apply $K$-means again to finding clusters, where we set $K = 5$. Then for each cluster, we find a set of users satisfying the $\gamma$-cluster criterion with the maximal size as a group-truth cluster.

\para{Evaluation Setting.} Each algorithm returns multiple sets of users in the end. For each ground-truth cluster, we pick a set with the highest F1 score. Then, we average the F1 scores of all picked sets and consider it as the accuracy of an algorithm.
This measurement reflects how well a cluster can be recovered, which has been widely used in clustering \cite{kloster2014heat, yin2017local}. 
For CLUB and SCLUB, the UCB parameter $\alpha$ is set as $\{0.8, 1.0, 1.2\}$.
For \sysn, the UCB of $\ntheta$ is set as 
$
\frac{ \sigma \sqrt{2d \log t + 2\log(2/\delta')}+1}{\sqrt{1+ m_{i,t}/4} \cdot n^{1/3} },
$
which is the similar form with Lemma \ref{lemma:ucb} but with the faster convergence rate.
We set $\gamma =0.2, \tau = \{8, 10, 12\} $ and $|S| = |N| = 100$. Next, we report the average accuracy of each methods.

%\vspace{-2em}
\begin{table}[t]
	\caption{Accuracy of $\gamma$-cluster detection on Synthetic and Yelp datasets.}\label{tab:1}
	\centering
	\begin{tabular}{c|ccc|ccc}
		\toprule
		     & \multicolumn{3}{|c|}{Synthetic} & \multicolumn{3}{|c}{Yelp}    \\
		     \midrule  
		     & F1  & Pre  & Recall  & F1  & Pre  & Recall \\
		\midrule  
		N-CLUB &0.390  & 0.246  &0.943 &0.484  & 0.334  &0.884  \\
		ST-CLUB  &0.578  & 0.549  &0.612 &0.626  & 0.593  &0.663 \\
		ST-SCLUB  &0.714 &0.745 & 0.687   &0.768 &0.863 & 0.693   \\
	    N-\sysn &0.662  & 0.618  &0.714 &0.675  & 0.620  &0.743\\
			\midrule  
		\sysn  & \textbf{0.880}  & \textbf{0.913} & \textbf{0.856} & \textbf{0.879}  & \textbf{0.908} & \textbf{0.853} \\

		\bottomrule
	\end{tabular}
%	\vspace{-1em}
\end{table}

\begin{table}[t]
	\caption{Accuracy of $\gamma$-cluster detection on MovieLens and Yahoo datasets.}\label{tab:2}
	\centering
	\begin{tabular}{c|ccc|ccc}
		\toprule
		     & \multicolumn{3}{|c|}{MovieLens} & \multicolumn{3}{|c}{Yahoo}   \\
		     \midrule  
		     & F1  & Pre  & Recall  & F1  & Pre  & Recall   \\
		\midrule  
		N-CLUB &0.417  & 0.286  &0.773 &0.454  & 0.334  &0.709  \\
		ST-CLUB  &0.520  & 0.429  &0.663 &0.528  & 0.385  &0.841 \\
		ST-SCLUB  &0.538  & 0.739  &0.424 &0.632  & 0.781  &0.532 \\
	    N-\sysn &0.472  & 0.432  &0.524 &0.615  & 0.553  &0.692\\
			\midrule  
		\sysn  & \textbf{0.814}  & \textbf{0.892} & \textbf{0.749} & \textbf{0.869}  & \textbf{0.935} & \textbf{0.813} \\
		\bottomrule
	\end{tabular}
	%\vspace{-1em}
\end{table}

Table \ref{tab:1} and Table \ref{tab:2} show \sysn's accuracy of cluster detection in comparison with four baselines. In general, \sysn significantly outperforms all the others. It starts with the seed users and explores their associated neighbors. With the proposed clustering strategy and termination criterion, \sysn achieves the ideal precision and recall. With the heuristic termination criterion, the clusters detected by N-CLUB are usually extremely large or small, containing a large number of false-positives or false-negatives. Even with the same number of rounds as \sysn, ST-CLUB still cannot reach the desired accuracy because it uses the connected components of a graph to represent clusters, which is influenced significantly by the edges created among user pairs. ST-SCLUB uses the top-down hierarchical clustering while it splits the users into small clusters very quickly . Its performance is better than ST-CLUB while still much worse than \sysn.
When N-\sysn terminates in a heuristic way, it either is still exploring the cluster structure or has over-explored, but it still achieves better performance than N-CLUB.

\hide{
\textit{Remark}. A good seed for \sysn should be the center of a ground-truth cluster, since it regards each seed as a center and clusters based on the center. With more seeds, the higher chance it has to achieve good performance. Due to the space limit, we report the variation of accuracy of \sysn with the change of the number of seeds in Supplementary. Empirically, with 20 seeds for 200 users, \sysn can achieve satisfactory performance.}

\begin{figure}[t] 
    \includegraphics[width=0.7\columnwidth]{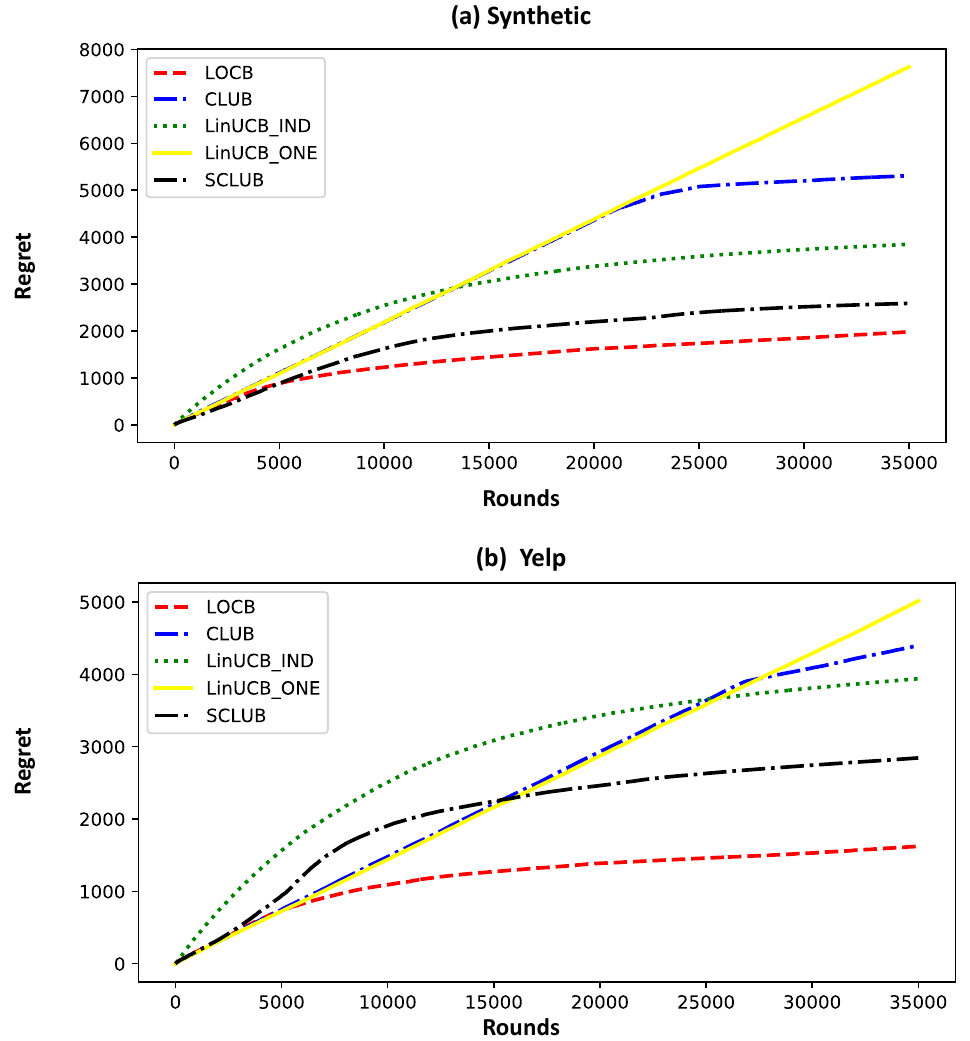}
    \centering
    \caption{ Regret comparison (\sysn) on Synthetic and Yelp datasets. }
       \label{fig:1}
\end{figure}
	%\vspace{-1em}

\begin{figure}[t] 
    \includegraphics[width=0.7\columnwidth]{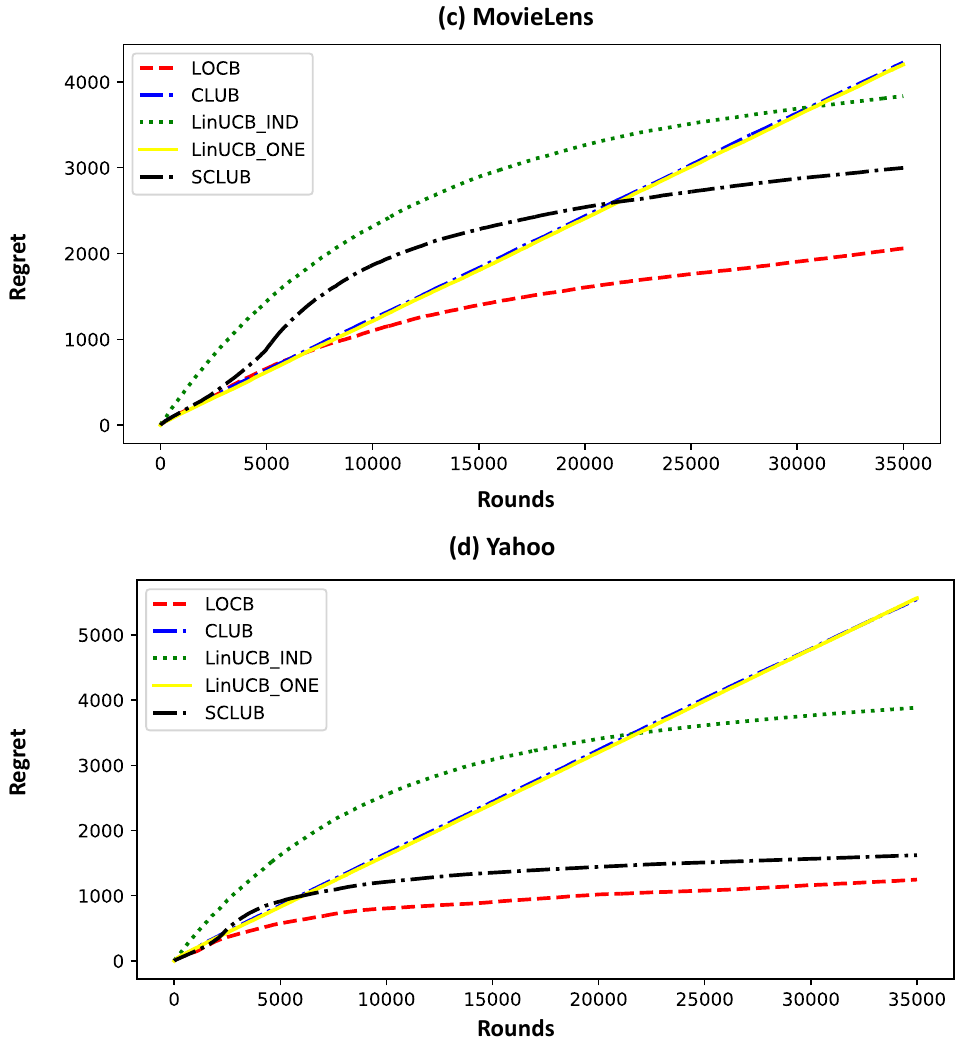}
    \centering
    \caption{ Regret comparison (\sysn) on MovieLens and Yahoo datasets.}
       \label{fig:2}
\end{figure}

\begin{figure}[t] 
    \includegraphics[width=0.66\columnwidth]{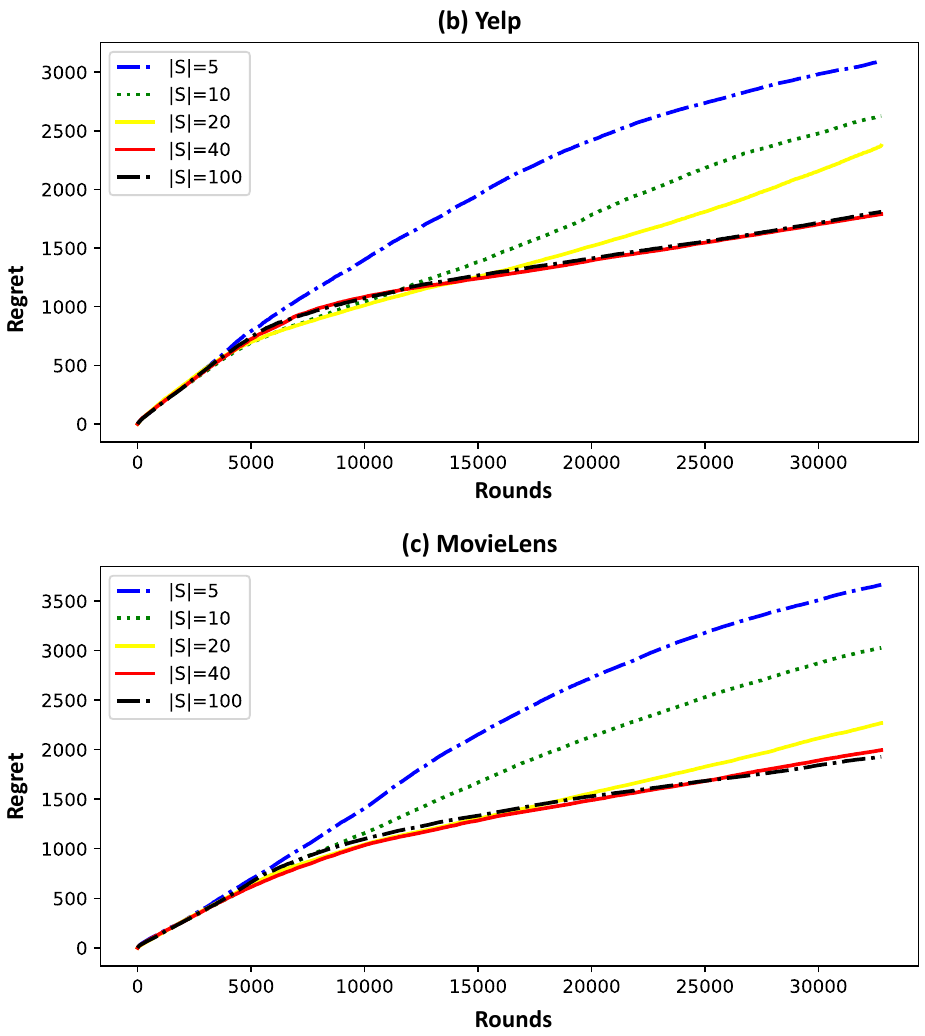}
    \centering
    \caption{The variation of regret of \sysn as the number of seeds increases on Yelp and MovieLens datasets.}
       \label{fig:seed}
\end{figure}

%\vspace{-1em}
\subsection{Regret Comparison.}
%\vspace{-1em}

In this sub-section, we evaluate the cumulative regret of \sysn compared to strong baselines.
Following \cite{2014onlinecluster, 2019improved}, the UCB of $\ntheta$ is set as $\sqrt{ \frac{ 1+ \log (1+t)}{1+t}}$ for all algorithms to accelerate the convergence rate.
For \sysn, we set $\gamma =0.2$ and seeds $|S| = 30 $ to consider the running time cost.
We choose four baselines: 
\begin{enumerate}
    \item linear bandit algorithms: LinUCB-ONE~\cite{2010contextual}, where all the users share one bandit parameters.  LinUCB-IND~\cite{2010contextual}, where each user has a separate bandit parameter.
    \item clustering of bandit algorithms: CLUB \cite{2014onlinecluster} and SCLUB~\cite{2019improved}.
\end{enumerate}

Figure \ref{fig:1} and Figure \ref{fig:2} show the cumulative regret of all methods on four datasets. 
As we can see, \sysn achieves the best performance compared to all baselines. With enough seeds, the Clustering module is able to recover each cluster effectively. Although the provided clusters are overlapping, the pulling module can find the best cluster for the observed user in each round. The performance of \sysn demonstrates that utilizing accurate user dependency can boost the performance of a bandit algorithm. 

For other baselines, in the beginning, the users' feedback is not enough to make an accurate estimate of $\ntheta_i$, and thus LinUCB-IND (IND) accumulates more regret than LinUCB-ONE (ONE). 
However, as more rounds are played, IND learns each user's preference more accurately, while ONE still uses one parameter for all users. 
Thus, ONE is outperformed by IND in the long run. As CLUB considers all the users as a cluster in the early phase, its performance is similar to ONE. 
As it learns the clusters progressively, its performance becomes better than ONE. 
However, the clusters CLUB finds are not accurate, resulting in much worse performance than \sysn.
SCLUB uses the set to represent each cluster and merges or splits them accordingly. 
However, SCLUB only starts clustering from one cluster center, making it incompetent to recover multiple clusters. 
Since it has higher accuracy than CLUB for finding good clusters, it performs better than CLUB but is still worse than \sysn.

\begin{table}[t] 
	\caption{The variation of clustering accuracy of \sysn as the number of seeds increases on Yelp and MovieLens datasets.}
	\centering
	\begin{tabular}{c|ccccc} 
		\toprule
		       & $|S|$ = 5  & $|S|$ = 10  & $|S|$=20  & $|S|$=50  & $|S|$ = 100\\
		\midrule  
		Yelp  & 0.457  & 0.733 & 0.823  & 0.853  &0.916  \\
		MovieLens  & 0.304  & 0.550  &0.747  & 0.832  &0.895 \\
		\bottomrule
	\end{tabular} \label{table:seed}
%	\vspace{-1em}
\end{table}

\begin{table}[t] 
	\caption{The variation of clustering accuracy of \sysn with different $\gamma$ on Yelp and MovieLens datasets.}
	\centering
	\begin{tabular}{c|ccccc} 
		\toprule
		       & $\gamma$ = 0.4  & $\gamma$ = 0.3  & $\gamma$=0.2  & $\gamma$=0.15  & $\gamma$ = 0.1\\
		\midrule  
		Yelp  & 0.732  & 0.908 & 0.916  & 0.812  &0.708  \\
		MovieLens  & 0.654  & 0.864  &0.895  & 0.762  &0.643 \\
		\bottomrule
	\end{tabular} \label{table:gamma}
%	\vspace{-1em}
\end{table}

\subsection{Effects of Parameters}

In this sub-section, we evaluate the effect of the two input parameters for \sysn, $S$ and $\gamma$.

Table \ref{table:seed} describes the variation of the accuracy of \sysn for cluster detection, as the number of seeds varies from [5, 100]. We use two datasets,  Yelp and MovieLens,  as representatives. When the number of seeds is smaller, the performance of \sysn boosts as it increases. Because with more seeds, the more chance \sysn has to find good seeds. 
For example, when $|S| =5$, it only can recover 2 out of 5 clusters; when $|S| = 20$, it almost recovers the total five clusters.      
A good seed for \sysn usually is the center of a ground-truth cluster. 
%While the number of seeds becomes larger, the performance of \sysn is more closed to the bottleneck. 

Figure \ref{fig:seed} reports the variation of regret of \sysn, as the number of seed increases. In accordance with the variation of accuracy of clustering, the performance of \sysn improves with the increasing of the number of seeds. Because the more seeds the Clustering module uses, the more accurate groups it finds. Given the candidate groups, the pulling module can find the optimal group for each user, and thus further decrease the regret. Similarly, when $|S|$ becomes large, the improvement becomes small.  Empirically, for 100 users, with more than 20 seeds, \sysn can achieve the ideal performance.

Table \ref{table:gamma} shows the change of clustering accuracy for \sysn with the varying $\gamma$. When $\gamma$ is a larger number (e.g., $\gamma=0.4$), the Clustering module will terminate earlier while with a low precision because each user lacks historical rewards and the confidence interval still is very large.
In contrast, when $\gamma$ is a smaller number (e.g.,$\gamma = 0.1$), the Clustering module will terminate much later to find $\gamma = 0.1$-clusters, where the returned clusters have higher precision but lower recall.
Therefore, the variance in accuracy of the gamma parameter becomes unimodal. 
The regret's change for \sysn with the varying of $\gamma$ also shows the unimodal shape, complying with change of clustering accuracy.
Due to the limited space, we will not show the regret change.

%\vspace{-1em}
\section{Conclusion}
%\vspace{-1em}

In this paper, we study the problem of detecting user clusters in contextual MAB. We propose \sysn, which utilizes a local procedure to cluster users and then leverages the best suitable cluster to improve the quality of recommendation for the serving user. In the theoretical analysis, we provide three theorems: (1) the returned set of users is a true cluster with probability at least $1-\delta$; (2) the termination of the Cluster module is bounded by  $O(n \log n)$; (3) the regret upper bound of \sysn is free of the number of seeds.  In the experiments, \sysn shows the promising empirical performance compared to strong baselines.

\section*{Acknowledgement}
This work is supported by National Science Foundation under Award No. IIS-1947203 and IIS-2002540. The views and conclusions are those of the authors and should not be interpreted as representing the official policies of the funding agencies or the government.
We would like to specially thank Zhiyong Wang for carefully checking the proofs of this paper, and Yunzhe Qi for the insightful discussions.

\bibliographystyle{ACM-Reference-Format}
\bibliography{ref.bib}

\section{Appendix}

The proof of \textbf{Lemma} \ref{lemma1} is as follows.

\begin{proof}
First, we define the complementary event of $\mathcal{E}$ as:
\[
\mathcal{E}' =  \left \{  \bigvee_{ \forall t \in [T], \forall i \in N}     \Arrowvert \hatt_{i, t} - \ntheta_{i} \Arrowvert >\bthe     \right \}.
\]
Then, 
\[
\begin{aligned}
&1 - \mathbb{P}(\mathcal{E} | I_T)  = \mathbb{P}(  \mathcal{E}' |I_T ) \\
& \leq  \sum_{i=1}^n  \left[\mathbb{P} \left( \forall t \in [T],  \Arrowvert \hatt_{i, t} - \ntheta_{i} \Arrowvert > \bthe \right)     \right]  \leq  \sum_{i=1}^n  \delta'  = \delta \\
&\Rightarrow  \mathbb{P}(\mathcal{E} | I_T) \geq 1-\delta.
\end{aligned}
\]
\end{proof}

The proof of \textbf{Theorem} \ref{theo:1} is as follows.

\begin{proof}
First, consider a cluster with respect to the seed $s \in S$.
With the probability $1- \delta$, suppose the event $\mathcal{E}$ happens. For any $i \in N$,  the the event $\mathcal{E}_{i, s}$ must happen, where
\[
\mathcal{E}_{i,s} = \left \{ \forall t,   \Arrowvert \hatt_{i,t} - \ntheta_i \Arrowvert \leq   \bthe   \bigwedge  \Arrowvert \hatt_{s,t} - \ntheta_s \Arrowvert \leq   \bthes \right \}
\]

Then, at any round $t$,  it has:
\[
\begin{aligned}
\Arrowvert \ntheta_{s} - \ntheta_{i} \Arrowvert &  =   \Arrowvert  \ntheta_{s} - \hatt_{s, t} + \hatt_{s, t} -  \hatt_{i, t} + \hatt_{i, t} - \ntheta_{i}  \Arrowvert \\
& \leq  \Arrowvert  \ntheta_{s} - \hatt_{s, t} \Arrowvert + \Arrowvert  \hatt_{s, t} -  \hatt_{i, t} \Arrowvert + \Arrowvert \hatt_{i, t} - \ntheta_{i}  \Arrowvert \\
& \leq     B_{\ntheta, s}(m_{s,t}, t) +  \Arrowvert  \hatt_{s, t} -  \hatt_{i, t} \Arrowvert +  \bthe.
\end{aligned}
\]
Let $\nst$ be the cluster output by Algorithm \ref{alg:main}.
For any round $t$, in Algorithm \ref{alg:main}, we remove $i$ from $\nst$ if $ \Arrowvert \hatt_{i, t} - \hatt_{s, t} \Arrowvert > \bthe  + \bthes, \forall t$. Thus, it indicates $\hat{\mathcal{E}}_{i, s}$ must happen, where 
\[
 \hat{\mathcal{E}}_{i, s} = \left \{ \forall t,   \Arrowvert \hatt_{i,t} -  \hatt_{s,t} \Arrowvert \leq  \bthe   +  B_{\ntheta, s}(m_{s,t}, t) \right \}
\]
Due to $\nst \subseteq N$, for any $i \in \mathcal{M}_{s}$, based on the inequations above,  we have 
\[
\Arrowvert \ntheta_{s} - \ntheta_{i} \Arrowvert   \leq 2 \left(  \bthe +  \bthes \right).
\]
Let $T$ be the number of rounds when the Clustering module terminates.
Then, according to Algorithm \ref{alg:main}, in round $T$, when it terminates, we have:
\[
\sup \{\bthe : i \in \mathcal{N}_{s,t}\} < \frac{\gamma}{8}. 
\]
Thus,
\[
\Arrowvert \ntheta_{s} - \ntheta_{i} \Arrowvert   \leq 2 \left(  B_{\ntheta, s}(m_{s,T}, T) +   B_{\ntheta, i}(m_{i,T}, T) \right) < \frac{\gamma}{2}.
\]

Thus, we have 
\begin{equation} \label{eq:7}
\forall i \in \nst,  i \not = s,   \Arrowvert \ntheta_{s} - \ntheta_{i} \Arrowvert <  \frac{\gamma}{2}.  
\end{equation}
Therefore, it holds 
\begin{equation} \label{eq:8}
\begin{aligned}
\forall i, j \in \nst,  i \not = s, j \not = s,     & \Arrowvert \ntheta_i   - \ntheta_{j}  \Arrowvert  =  \Arrowvert  \ntheta_i - \ntheta_{s} + \ntheta_{s} - \ntheta_{j}        \Arrowvert \\
&\leq  \Arrowvert  \ntheta_i - \ntheta_{s} \Arrowvert + \Arrowvert   \ntheta_{s} - \ntheta_{j}  \Arrowvert < \gamma. 
\end{aligned}
\end{equation}
Putting Eq (\ref{eq:7}) and Eq (\ref{eq:8}) together, this directly proves that the return set $\nst$ by \sysn is a true $\gamma$-cluster.
Similarly, for each $s \in S$, $\nst$ is a true $\gamma$-cluster,
with probability $1-\delta$. 
\end{proof}

The proof of \textbf{Theorem} \ref{theo:2} is as follows.

\begin{proof}
Let $\bthes$ be the UCB in Lemma \ref{lemma:ucb}.
Based on Lemma 7 in \cite{2014onlinecluster}, it has
\[
\bthes
\leq  \frac{\sigma \sqrt{2d \log(1+t) + 2\log(2/\delta') }+1 }{\sqrt{1 + \lambda m_{s,t} /8 }}
\]
when 
\[
m_{s,t} \geq \frac{2 \times 32^2 }{\lambda^2} \log \left ( \frac{2nd}{\delta'} \right) \log \left ( \frac{32^2}{\lambda^2} \log\left( \frac{2nd}{\delta'} \right) \right). 
\]
Let an iteration represent $n$ rounds. Then, let $T'$ be the number of rounds when the Clustering module terminates and $T = T' - n$.
According to Algorithm \ref{alg:main}, before the last iteration, give a seed $s$, it has
\[
\begin{aligned}
\frac{\gamma}{8} & \leq   \bthes \\
 & \leq  \frac{\sigma \sqrt{2d \log(1+T) + 2\log(2/\delta') }+1 }{\sqrt{1 + \lambda m_{s,T} /8 }}  \\
& \leq   \frac{\sigma \sqrt{2d \log(1+T) + 2\log(2/\delta') }}{\sqrt{1 + \lambda m_{s,T} /8 }}  + \frac{1}{\sqrt{1 + \lambda m_{s,T} /8 }}\\
\frac{\gamma}{16}  & \leq \frac{1}{2}\left[  \frac{\sigma \sqrt{2d \log(1+T) + 2\log(2/\delta') }}{\sqrt{1 + \lambda m_{s,T} /8 }}  + \frac{1}{\sqrt{1 + \lambda m_{s,T} /8 }}  \right]
\end{aligned}
\]
According to Jensen's inequality, we have
\[
\begin{aligned}
\frac{\gamma^2}{16^2} & \leq \frac{\sigma^2 \left( 2d \log(1+T) + 2\log(2/\delta') \right) }{1 + \lambda m_{s,T} /8 } + \frac{1}{1 + \lambda m_{s,T} /8} \\ 
(1 + \lambda m_{s,T} /8) \frac{\gamma^2}{16^2} & \leq \sigma^2 \left( 2d \log(1+T) + 2\log(2/\delta') \right) +1 \\
\frac{\gamma^2}{16^2} & \leq  \sigma^2 \left( 2d \log(1+T) + 2\log(2/\delta') \right) +1 -   \frac{ \lambda m_{s,T}\gamma^2 }{16^2\times8}\\
\frac{\gamma^2}{512 \sigma^2 } & \leq    d \log(1+T) + \log(2/\delta')  + \frac{1}{2\sigma^2} -   \frac{ \lambda m_{s,T}\gamma^2 }{16^3 \sigma^2}\\
\end{aligned}
\]
Replace $\frac{ \lambda \gamma^2 }{16^3 \sigma^2}$ by $C$ and $\delta'$ by $\frac{\delta}{n}$, then
\[
\begin{aligned}
\frac{\gamma^2}{512 \sigma^2 } - \frac{1}{2\sigma^2} & \leq  d \log(1+T) + \log(
\frac{2n}{\delta})  -  C m_{s,T}\\
\end{aligned}
\] 
As $T \geq 1 $, it holds that
\[
\begin{aligned}
\frac{\gamma^2}{512 \sigma^2 } - \frac{1}{2\sigma^2} & \leq d\log 2T  + \log(\frac{2n}{\delta})  -  C m_{s,T}  \\
& = d\log T + \log(\frac{2^{(d+1)}n}{\delta}) -  C m_{s,T} 
\end{aligned}
\]

Suppose each user is evenly served. Thus 
\[
 m_{s,T} = \frac{T}{n}.
\]
We have
\[
\begin{aligned}
\frac{\gamma^2}{512 \sigma^2 } - \frac{1}{2\sigma^2} & \leq    d\log T + \log(\frac{2^{(d+1)}n}{\delta}) - \frac{CT}{n} \\ 
 \log T & \geq  \frac{CT}{nd} + \frac{1}{d} \left( \frac{\gamma^2 - 256}{512\sigma^2} -  \log(\frac{2^{(d+1)}n}{\delta}) \right)
\end{aligned}
\]

According to Lemma \ref{lemma2}, we have 
\[
\begin{aligned}
T &<  \frac{2nd}{C} \left[ \log \frac{nd}{C} +   \frac{1}{d} \left(  \log(\frac{2^{(d+1)}n}{\delta}) - \frac{\gamma^2 - 256}{512\sigma^2} \right)   \right]\\
& =    \frac{2nd}{C} \log \frac{nd}{C}  +  \frac{2n}{C}\left(  \log(\frac{2^{(d+1)}n}{\delta}) - \frac{\gamma^2 - 256}{512\sigma^2} \right)  
\end{aligned} 
\]
Because $T = T' -n$, it has that 

\[
T' < \frac{2nd}{C} \log \frac{nd}{C}  +  \frac{2n}{C}\left(  \log(\frac{2^{(d+1)}n}{\delta}) - \frac{\gamma^2-256}{512\sigma^2} \right)  + n.
\]

This directly proves Theorem \ref{theo:2}.
\end{proof}

\begin{lemma} \label{lemma2}
Let a >0. For any  $ at + b \leq \log t,  t < (2/a)[\log(1/a) - b]$.
\end{lemma}
\begin{proof}
%This lemma is inferred from Lemma 7 in \cite{2010active}.

Let $f(t) = at + b $ and $h(t) =  \log t$. Assume $t_0$ satisfies $f'(t_0) = h'(t_0)$, and then $t_0 = \frac{1}{a}$. We have
\[
\begin{cases}
f'(t) \geq h'(t),  \ \text{if} \ t \geq t_{0}; \\
f'(t) < h'(t),  \ \text{otherwise}.  \\
\end{cases}
\]
As $ f(t) \leq h(t)$, it must have 
\begin{equation} \label{eq:1}
f(t_0) \leq h(t_0)  \Rightarrow \log \frac{1}{a} -b \geq 1.
\end{equation}
Because if $f(t_0) > h(t_0)$,  for $t > t_0$ , it has  $f'(t) > h'(t)$,  $f(t) > h(t)$; for $t < t_0$,  it has $f'(t) < h'(t), f(t) > h(t)$. This is a contradiction. Thus, Eq (\ref{eq:1}) is true.

Let $t_1  = (2/a)[\log(1/a) - b]$, and replace $t$ by $t_1$, we have
\[
\begin{cases}
&f(t_1) = \log \frac{1}{a} +  \log \frac{1}{a} - b \\
&h(t_1) = \log \frac{2}{a} + \log \left(\log \frac{1}{a} - b \right).\\
\end{cases}
\]
Because $\log\frac{1}{a} -b \geq 1$, we have
\[ 
f(t_1) > h(t_1).
\] 
As $t_1 > t_0$, for any $t \geq t_1$, we have   $f'(t_0) > h'(t_0)$, $f(t)>h(t)$. Therefore, for  $f(t) \leq h(t)$, i.e., $ at + b \leq \log t$, we have 
\[
t <  t_1 = (2/a)[\log(1/a) - b].
\] 

\end{proof}

\begin{lemma} \label{lemma3}
Given a user $i \in N$ and a context vector $\bx_t$, define 
\[
\mathbb{P}( \forall t \in [T],  | \ntheta_{i}^{\intercal} \bx_t - \hatt_{i,t}^{\intercal} \bx_t | > \cbri ) < \delta',
\] 
where $\delta'$ is a confidence level with respect to $\delta$ and $t$.
Define the random event 
\[
\mathcal{F} =  \left \{  \bigwedge_{\forall t \in [T], \forall i \in N}   | \ntheta_{i}^{\intercal} \bx_t - \hatt_{i,t}^{\intercal} \bx_t | \leq \cbri \right \}
\]
Let  $I_T = [i_1, i_2, \dots, i_T]$ be the sequence of served users in each round up to $T$. If $\delta'$ is defined as
\[
\delta' = \frac{ \delta}{n},
\]
then, the probability of $\mathcal{F}$ happening is at least $1-\delta$,
\[ 
i.e., \mathbb{P}(\mathcal{F} | I_T) >  1- \delta.
\]
\end{lemma}
\begin{proof}
The proof is similar to the proof of Lemma \ref{lemma1}.
\end{proof}

\begin{lemma} \label{lemma4}
Let $\nst \subseteq N $ be a cluster with respect to seed $s$. The bandit parameter of $\nst$ is defined as  $\tnst =  \frac{1}{|\nst|} \sum_{i \in \nst} \ntheta_{i}$ and its estimation is defined as  $\htnst =  \frac{1}{|\nst|} \sum_{i \in \nst} \hatt_{i,t}$. Then, with probability $1-\delta$, $\forall t, \forall s \in S$, it has
\[
|\tnst^{\intercal} \bx_t - \htnst^{\intercal} \bx_t | \leq \cbnst,
\]
where $\cbnst  = \frac{1}{|\nst|} \sum_{i \in \nst} \cbri$.
\end{lemma}

\begin{proof}
Based on Lemma \ref{lemma3}, with probability $1-\delta$, the event $\mathcal{F}$ happens. Then, $\forall t, \forall s \in S$, it has
\[
\begin{aligned}
|\tnst^{\intercal} \bx_t - \htnst^{\intercal} \bx_t |  &  =
\left| \frac{1}{|\nst|} \sum_{i \in \nst} \ntheta_{i}^{\intercal} \bx_t  - \frac{1}{|\nst|} \sum_{i \in \nst} \hatt_{i,t}^{\intercal}  \bx_t  \right |  \\
& =  \frac{1}{|\nst|} \sum_{i \in \nst} \left | \ntheta_{i}^{\intercal} \bx_t -  \hatt_{i,t}^{\intercal}  \bx_t \right| \\
&  \leq \frac{1}{|\nst|} \sum_{i \in \nst} \cbri = \cbnst,
\end{aligned}
\]
as claimed.

\end{proof}

The proof of \textbf{Theorem} \ref{tho:bound} is as follows.

\begin{proof}
The regret of round $t$ is defined as
$$
\begin{aligned}
R_t & =  \ntheta_{i}^{\intercal}  \bx_t^{*}  - \ntheta_{i}^{\intercal}  \bx_{t} \\
& = \ntheta_{i}^{\intercal}  \bx_t^{*}  -  \tnst^{\intercal}\bx_t^{*} +  \tnst^{\intercal}\bx_t^{*}  -  \htnst^{\intercal}\bx_t^{*} +  \htnst^{\intercal}\bx_t^{*}  -  \ntheta_{i}^{\intercal}  \bx_{t}\\
& \leq  |\ntheta_{i}^{\intercal}  \bx_t^{*}  -  \tnst^{\intercal}\bx_t^{*}| +  |\tnst^{\intercal}\bx_t^{*}  -  \htnst^{\intercal}\bx_t^{*}| +  \htnst^{\intercal}\bx_t^{*}  -  \ntheta_{i}^{\intercal}  \bx_{t}.
\end{aligned}
$$
Based on the $UCB$, $| \tnst^{\intercal}\bx_t^{*}  -  \htnst^{\intercal}\bx_t^{*} | \leq \cbnst^* $,
it has
\[
R_t  \leq  |\ntheta_{i}^{\intercal}  \bx_t^{*}  -  \tnst^{\intercal}\bx_t^{*}| + \cbnst^* +   \htnst^{\intercal}\bx_t^{*}  -  \ntheta_{i}^{\intercal}  \bx_{t}.
\]

Let $\nst$ be the cluster that $i$ belongs to and $\xt$ be the selected context vector in $t$.
Based on \sysn, it has: 
\[
\bx_{t} = \arg  \max_{\bx_{a,t} \in \mathbf{X}_t} \max_{s \in S(i)} \left( \htnst^{\intercal} \bx_{a,t} + \cbnst \right).  
\]
Thus, we have  
\[ 
\cbnst^* +   \htnst^{\intercal}\bx_t^{*} \leq \cbnst  +  \htnst^{\intercal}\bx_t. 
\]
Then,
\[
R_t \leq  | \ntheta_{i}^{\intercal}  \bx_t^{*}  -  \tnst^{\intercal}\bx_t^{*} |+ \cbnst  + | \htnst^{\intercal}\bx_t  -  \ntheta_{i}^{\intercal}  \bx_{t}|.
\] 
As
\[
\begin{aligned}
| \htnst^{\intercal}\bx_t  -  \ntheta_{i}^{\intercal}  \bx_{t} | &=  |\htnst^{\intercal}\bx_t - \tnst ^{\intercal}\bx_t +  \tnst ^{\intercal}\bx_t -  \ntheta_{i}^{\intercal}  \bx_{t} | \\ 
&\leq  |\htnst^{\intercal}\bx_t - \tnst ^{\intercal}\bx_t| +  |\tnst ^{\intercal}\bx_t -  \ntheta_{i}^{\intercal}  \bx_{t} | \\
& \leq \cbnst  + | \tnst ^{\intercal}\bx_t -  \ntheta_{i}^{\intercal}  \bx_{t} |,
\end{aligned}
\] 
we have 
\[
R_t  \leq  |\ntheta_{i}^{\intercal}  \bx_t^{*}  -  \tnst^{\intercal}\bx_t^{*}| +  | \tnst ^{\intercal}\bx_t -  \ntheta_{i}^{\intercal}  \bx_{t} | + 2 \cbnst .
\]
The accumulated regret of $T$ rounds is defined as:
\[
\mathbf{R}_T = \sum_{t=1}^T R_t \leq \sum_{t=1}^T \left(  |\ntheta_{i}^{\intercal}  \bx_t^{*}  -  \tnst^{\intercal}\bx_t^{*}| +  | \tnst ^{\intercal}\bx_t -  \ntheta_{i}^{\intercal}  \bx_{t} | \right)  +   \sum_{t=1}^T 2 \cbnst.
\]
Thus, $\mathbf{R}_T$ is determined by two items. First, let us bound the second item. 
\[
\begin{aligned}
\sum_{t=1}^{T}  \cbnst   & = \sum_{t=1}^{T}  \frac{1}{\nst} \sum_{i \in \nst} \cbri \\
\end{aligned}
\] 
For each $i \in \nst$, it has
\[
\cbri =  \alpha \sqrt{\bx_t^{\intercal} \mathbf{A}_{i, t}^{-1} \bx_t} =  \alpha \|\mathbf{x}_t \|_{\mathbf{A}_{i, t}^{-1}} 
\]
Recall the Theorem 3 in ~\cite{2011improved}, $\alpha = O\left(\sqrt{d\log \frac{m_{i,t}}{\delta}}\right)$.
Then, according to Lemma \ref{lemma3}, replace $\delta$ by $\delta'$, to make sure $\mathcal{F}$ happens with probability at least $1-\delta$. It has
\[
\alpha = O \left( \sqrt{ d  \log  \left( m_{i,t} n/\delta  \right ) }  \right)
\]
Thus,
\[
\begin{aligned}
\forall i \in \nst,
\sum_{t=1}^{T}  \cbri & = O \left( \sqrt{ d  \log  \left( m_{i,t} n/\delta  \right ) }  \right) \cdot \sum_{t=1}^{T} \|\mathbf{x}_t \|_{\mathbf{A}_{i, t}^{-1}}  \\
 & \leq O \left( \sqrt{ d  \log  \left(  m_{i,T}  n/\delta  \right ) }  \right) \cdot 
 \sqrt{ T \sum_{t=1}^T \|\mathbf{x}_t \|_{\mathbf{A}_{i, t}^{-1}}^2}\\
 &  \leq O \left( \sqrt{ d  \log  \left(m_{i,T}   n/\delta  \right ) }  \right) \cdot 
 \sqrt{ 2 T  n  \log \left( \frac{  \text{det}(\mathbf{A}_{i, T})}{ \text{det}(\lambda \mathbf{I})  }  \right)  }\\
& \leq \sqrt{Tn}  \cdot
\sqrt {2d\log (1+T/dn) } \cdot O \left( \sqrt{ d  \log  \left(T/\delta  \right ) }  \right) 
\end{aligned}
\] 
where the second inequality is based on the Lemma 11 in \cite{2011improved} and the last inequality is based on the Lemma 10 in \cite{2011improved} and replacing $m_{i,T}$ by $T/n$.
Therefore,
\[
\begin{aligned}
\sum_{t=1}^{T}  \cbnst & = \frac{1}{\nst} \sum_{i \in \nst} O \left( \sqrt{ d  \log  \left(m_{i,T} n/\delta  \right ) }  \right) \cdot   \sum_{t=1}^{T}  \|\mathbf{x}_t \|_{\mathbf{A}_{i, t}^{-1}} \\
& \leq   \sqrt{Tn}  \cdot
\sqrt {2d\log (1+T/dn) } \cdot O \left( \sqrt{ d  \log  \left(T/\delta  \right ) }  \right).
\end{aligned}
\]

Second, let us bound the first item. We need to consider two condition. First, suppose $\nst$ is a true $\gamma$-cluster and it has

\[
\begin{aligned}
| \tnst^{\intercal}\bx_t - \ntheta_{i}^{\intercal}  \bx_t | & = \left| \frac{1}{|\nst|}  \sum_{j \in \nst} \ntheta_{j}^{\intercal} \bx_t  -   \ntheta_{i}^{\intercal}  \bx_t \right| \\
 & = \frac{1}{|\nst|}  \sum_{j \in \nst}  \left| \ntheta_{j}^{\intercal}  \bx_t - \ntheta_{i}^{\intercal}  \bx_t \right| \\
 & = \frac{1}{|\nst|}  \sum_{j \in \nst} \left| \bx_t^{\intercal}  ( \ntheta_{j}   - \ntheta_{i})   \right|
\end{aligned}
\]
According to Cauchy-Shwartz inequality,  
\[
\begin{aligned}
| \tnst^{\intercal}\bx_t - \ntheta_{i}^{\intercal}  \bx_t | & \leq  \frac{1}{|\nst|}  \sum_{j \in \nst} \|\xt  \| \cdot \| \ntheta_{j}   - \ntheta_{i} \| \\
& \leq  \frac{1}{|\nst|}  \sum_{j \in \nst} \| \ntheta_{j}  - \ntheta_{i} \|  \\
& \leq \frac{1}{|\nst|}  \sum_{j \in \nst} \gamma = \gamma,
\end{aligned}
\]
if  $\nst$ is a  $\gamma$-cluster.

Now, let us consider the condition that $\nst$ is not a $\gamma$-cluster. 
Based on the proof of Theorem \ref{theo:1} ($ \bm{i} = \bthe$), it has 
\[
\| \ntheta_{j}  - \ntheta_{i} \| \leq  2\left(  \bm{i} +  \bm{j} \right).
\]
Therefore,
\[
\begin{aligned}
| \tnst^{\intercal}\bx_t - \ntheta_{i}^{\intercal}  \bx_t | & \leq \frac{1}{|\nst|}  \sum_{j \in \nst} \Arrowvert \ntheta_{j}  - \ntheta_{i} \Arrowvert \\
& \leq \frac{1}{|\nst|}  \sum_{j \in \nst} 2\left(  \bm{i} +  \bm{j} \right) \\
\end{aligned}
\] 
With UCB of Lemma ~\ref{lemma1}, it has 
\[
\begin{aligned}
\forall i \in \nst,   \bm{i} & \leq  \sigma \sqrt{2d \log T + 2\log(2/\delta')}+1\\
&\leq O \left( \sqrt{ d  \log  \left(Tn/\delta  \right ) }  \right) = A 
\end{aligned}
\]
Let $\bar{U}$ be the upper bound in Theorem \ref{theo:2}. Then, based on Theorem \ref{theo:2}, with probability at least $1-\delta$, for each $s \in S$, $\nst$ is a $\gamma$-cluster when $T \geq \bar{U}$.
Therefore, for the first item of $\mathbf{R}_T$, we have
\[
\begin{aligned}
 \sum_{t=1}^T \left(  |\ntheta_{i}^{\intercal}  \bx_t^{*}  -  \tnst^{\intercal}\bx_t^{*}| +  | \tnst ^{\intercal}\bx_t -  \ntheta_{i}^{\intercal}  \bx_{t} | \right)  \leq \sum_{t=1}^{\bar{U}} 4 A +  \sum_{t = \bar{U}}^T 2 \gamma \\
 = O\left(nd \log nd \right) \cdot O \left( \sqrt{ d  \log  \left(Tn/\delta  \right ) }  \right) + \left(T - O\left(nd \log nd \right)\right) \gamma,
 \end{aligned}
\]
because  $\bar{U} = O\left(nd \log nd \right) $.
Then, putting these two items together, we have
\[
\begin{aligned}
\mathbf{R}_T & \leq   O\left(nd \log nd \right) \cdot O \left( \sqrt{ d  \log  \left(Tn/\delta  \right ) }  \right) + \left(T - O\left(nd \log nd \right)\right) \gamma \\
&+ \sqrt{Tn}  \cdot
\sqrt {2d\log (1+T/dn) } \cdot O \left( \sqrt{ d  \log  \left(T/\delta  \right ) }  \right),
\end{aligned}
\]
which proves the claim.
\end{proof}

\end{document}